\documentclass{article}

\usepackage[preprint,nonatbib]{neurips_2025}

\usepackage[utf8]{inputenc} 
\usepackage[T1]{fontenc}    
\usepackage{hyperref}       
\usepackage{url}            
\usepackage{booktabs}       
\usepackage{amsfonts}       
\usepackage{nicefrac}       
\usepackage{microtype}      
\usepackage{xcolor}         

\usepackage{graphicx} 
\usepackage{multirow}
\usepackage{multicol}
\usepackage{booktabs}
\usepackage{pifont}
\usepackage{enumitem}
\usepackage{bm}
\usepackage{amsmath}

\newcommand{\xmark}{\ding{55}}%

\title{OB3D: A New Dataset for Benchmarking Omnidirectional 3D Reconstruction Using Blender}

\author{%
  \textbf{Shintaro Ito}$^1$, \textbf{Natsuki Takama}$^1$, \textbf{Toshiki Watanabe}$^1$, \textbf{Koichi Ito}$^1$,\\
  \textbf{Hwann-Tzong Chen}$^2$, \textbf{and} \textbf{Takafumi Aoki}$^1$\\
  $^1$ Tohoku University \ \ \ $^2$ National Tsing Hua University\\
  \texttt{shintaro@aoki.ecei.tohoku.ac.jp}
}

\begin{document}

\maketitle

\begin{abstract}
  Recent advancements in radiance field rendering, exemplified by Neural Radiance Fields (NeRF) and 3D Gaussian Splatting (3DGS), have significantly progressed 3D modeling and reconstruction.
  The use of multiple 360-degree omnidirectional images for these tasks is increasingly favored due to advantages in data acquisition and comprehensive scene capture.
  However, the inherent geometric distortions in common omnidirectional representations, such as equirectangular projection (particularly severe in polar regions and varying with latitude), pose substantial challenges to achieving high-fidelity 3D reconstructions.
  Current datasets, while valuable, often lack the specific focus, scene composition, and ground truth granularity required to systematically benchmark and drive progress in overcoming these omnidirectional-specific challenges.
  To address this critical gap, we introduce Omnidirectional Blender 3D (OB3D), a new synthetic dataset curated for advancing 3D reconstruction from multiple omnidirectional images.
  OB3D features diverse and complex 3D scenes generated from Blender 3D projects, with a deliberate emphasis on challenging scenarios.
  The dataset provides comprehensive ground truth, including omnidirectional RGB images, precise omnidirectional camera parameters, and pixel-aligned equirectangular maps for depth and normals, alongside evaluation metrics.
  By offering a controlled yet challenging environment, OB3D aims to facilitate the rigorous evaluation of existing methods and prompt the development of new techniques to enhance the accuracy and reliability of 3D reconstruction from omnidirectional images.
\end{abstract}

\section{Introduction}
\label{sec:introduction}

Omnidirectional images, captured using an omnidirectional camera, play a crucial role in computer vision tasks aimed at recognizing and understanding 3D spaces.
Unlike perspective projection images captured using a standard camera, omnidirectional images provide a wider field of view, enabling the comprehensive capture of spatial information \cite{AiCW25}.
Among various computer vision tasks, 3D reconstruction is particularly significant, with broad-ranging use cases such as AR/VR environment generation, high-precision 3D mapping for autonomous driving, and cultural heritage digital archiving.
By leveraging omnidirectional images, these tasks can overcome the limitations imposed by the restricted field of view of standard cameras and facilitate the efficient reconstruction of wide-area indoor and outdoor spaces.

So far, there have been proposed datasets for real-world objects \cite{ChangDFHNSSZZ17,ArmeniSZS17,HuangCZY22,KimCJK21,CruzHLKBK21,BertelYLR20} and synthetic 3D objects \cite{ZioulisKZD18,ChoiK023,HuangY22,WangHCLYSCS19,SongYZCSF17} for recognizing and understanding 3D space using omnidirectional images.
These datasets include RGB images, 2D and 3D segmentation masks, depth maps, normal maps, and 3D mesh models of the scene, and can be used for room layout estimation \cite{WangYSCT21,SunTSCSTWC24,CruzHLKBK21}, depth map estimation \cite{ZioulisKZD18,ReyareaYR22}, object surface normal estimation \cite{HuangZD24}.
Many of these datasets are publicly available on the Internet and are widely utilized in the research on omnidirectional images.
With the recent rapid development of novel view synthesis techniques such as Neural Radiance Fields (NeRF) \cite{MildenhallSTBRN20} and 3D Gaussian Splatting (3DGS) \cite{KerblKLD23}, there have also been proposed omnidirectional image datasets \cite{HuangCZY22,ChoiK023} for the purpose of novel view synthesis.
These techniques integrate multiple view images to generate high-quality view transformations and rendering, thereby providing a more refined 3D representation of scenes.
The rendering of high-quality RGB images and depth maps using NeRF and 3DGS has a significant impact on 3D reconstruction from omnidirectional images.
3D reconstruction from omnidirectional images has faced the problem of correcting geometric distortion caused by the wide field of view.
By using NeRF and 3DGS, information from multiple views can be integrated, resulting in accurate shape estimation and improved consistency among views.
This approach enables high-fidelity estimation of 3D structures and improves the accuracy of large-scale scene reconstruction from multiple omnidirectional images.

On the other hand, existing datasets lack an evaluation environment specialized for 3D reconstruction from omnidirectional images.
Many datasets are mainly intended for object recognition and scene understanding, and lack the ground truth and evaluation protocols for quantitatively evaluating the accuracy of mesh models obtained through 3D reconstruction.
In addition, some datasets contain data with insufficient overlap between images, which is necessary for 3D reconstruction from multi-view images, making it impossible to properly evaluate the actual 3D reconstruction performance in some cases.
Another challenge is the bias in camera trajectories with existing datasets.
Many datasets are collected with an egocentric camera trajectory, i.e., the camera captures the surroundings with the photographer in the center, while actual capturing with an omnidirectional camera requires a non-egocentric trajectory, i.e., the camera moves freely within the scene while capturing the images.
This difference makes it difficult to adapt to the task of 3D reconstruction from real omnidirectional images.
Therefore, existing studies have created their own datasets for each experiment and evaluated each method \cite{JangMKKRK22,KimMJTK24,ChiuWSC23,ZhongmiaoQSJXRL24}.
The lack of a standardized benchmark makes it difficult to conduct fair comparative experiments.

In this paper, we propose a novel dataset called the Omnidirectional Blender 3D (OB3D) dataset for evaluating 3D reconstruction methods using omnidirectional images and a benchmark for evaluating accuracy.
OB3D includes 12 indoor and outdoor scenes, each of which provides RGB images, depth maps, normal maps, sparse 3D point clouds, and ground-truth camera parameters, allowing quantitative evaluation of the performance of 3D reconstruction methods.
This dataset can also be used for related tasks such as novel viewpoint synthesis and camera calibration, contributing to the development of 3D reconstruction technologies from omnidirectional images.

The following are features and contributions of OB3D.
\begin{itemize}
[leftmargin=8mm]
    \item So far, there have been proposed datasets containing omnidirectional images \cite{ChangDFHNSSZZ17,ArmeniSZS17,HuangCZY22,KimCJK21,CruzHLKBK21,BertelYLR20,ZioulisKZD18,ChoiK023,HuangY22}.
    On the other hand, there have been no datasets or benchmarks proposed specifically for evaluating the accuracy of 3D reconstruction.
    OB3D proposed in this paper provides omnidirectional images, camera parameters, depth maps, normal maps, and an evaluation protocol for 3D reconstruction, enabling fair and standardized accuracy evaluation against existing methods.
    \item In OB3D, the camera pose of each image is defined by the exact camera parameters generated by Blender\footnote{\url{http://www.blender.org}\label{fot:blender}}.
    When estimating the camera parameters for each viewpoint from real-world images, it is necessary to perform Simultaneous Localization and Mapping (SLAM) or Structure from Motion (SfM), which can introduce errors.
    The camera parameters provided by OB3D are more accurate than those estimated from real-world images and can be used for accurate evaluation of 3D reconstruction methods using omnidirectional images.
    \item Most of the existing synthetic datasets are image sets \cite{ChoiK023} with the egocentric camera trajectory or sparse image sets with sparse overlap between images \cite{ChangDFHNSSZZ17}.
    In practical situations, an omnidirectional camera is held in the hand while moving, or mounted on a drone to capture wide areas.
    Therefore, it is important to have an image set that includes not only the egocentric trajectory but also the non-egocentric trajectory, which captures the object from a free viewpoint, to evaluate the versatility of methods.
    Since OB3D includes both trajectories, we can evaluate the robustness of methods with respect to the camera trajectories.
    \item OB3D contains omnidirectional images, camera parameters, RGB images, depth and normal maps, as well as a sparse 3D point cloud reconstructed using open-source SfM software.
    These data can be used to evaluate novel viewpoint synthesis and camera parameter estimation methods.
    In this paper, we demonstrate the versatility of OB3D by conducting accuracy evaluation experiments for various tasks using OB3D.
\end{itemize}

The data of OB3D is available on Kaggle\footnote{\url{https://www.kaggle.com/datasets/shintacs/ob3d-dataset}\label{fot:kaggle}}, and its evaluation code is available on GitHub\footnote{\url{https://github.com/gsisaoki/Omnidirectional_Blender_3D_Dataset}\label{fot:github}}.

\section{Related Work}

We give a brief overview of topics related to omnidirectional images and datasets.

\subsection{Topics on Omnidirectional Images}

Multi-view 3D reconstruction can obtain highly accurate mesh models and 3D point clouds, and novel view synthesis can generate photorealistic images or 3D representations.
3D reconstruction using perspective projection images has been widely studied \cite{MildenhallSTBRN20,KerblKLD23,SchonbergerF16,SchonbergerZPF16,SDFStudio}, and recently, many methods utilizing omnidirectional images have been proposed.
Omnidirectional images are suitable for reconstructing a wide area of the scene since they overcome the viewing angle limitation of perspective projection images and can comprehensively acquire information on the entire space.
In the following, we describe the major tasks utilizing omnidirectional images, including camera parameter estimation, novel view synthesis, and 3D reconstruction.

\noindent
{\bf Camera Parameter Estimation} ---
The estimation of the position and intrinsic parameters of the camera from omnidirectional images primarily employs methods based on SLAM and SfM as in perspective projection images.
The software for omnidirectional images are available as OpenSLAM\footnote{\url{https://github.com/OpenSLAM}\label{fot:openslam}}, OpenMVG\footnote{\url{https://github.com/openMVG/openMVG}\label{fot:openmvg}}, OpenSfM\footnote{\url{https://opensfm.org/}\label{fot:opensfm}}, SphereSfM\footnote{\url{https://github.com/json87/SphereSfM}\label{fot:spheresfm}}.

\noindent
{\bf Novel View Synthesis} ---
With the advent of NeRF \cite{MildenhallSTBRN20} and 3DGS \cite{KerblKLD23}, novel view synthesis techniques that take advantage of multi-viewpoint images have been dramatically developed \cite{HuangYCGG24,BarronMVSH22,LiuHOM25,YuCHSG24}.
NeRF \cite{MildenhallSTBRN20} utilizes the rays corresponding to each image as training data to achieve novel view synthesis that takes into account the continuity between views.
On the other hand, 3DGS \cite{KerblKLD23} generates images at unknown views by placing 3D Gaussians on the world coordinate system and rasterizing them to the corresponding camera view.
With the advancement of novel view synthesis techniques, NeRF and 3DGS-based rendering that takes into account the camera model of a omnidirectional camera is now available, and methods such as 360-GS \cite{Jiayang24}, ODGS \cite{LeeCHL24}, op43dgs \cite{HuangBGLG24}, OmniGS \cite{LiHYC24}, EgoNeRF \cite{ChoiK023}, and PanoHDRNeRF \cite{GeraDRNL23} have been proposed.
An omnidirectional image has a wide field of view of horizontal 360 degrees and vertical 180 degrees, and can capture a wide range of space, making it suitable for novel view synthesis for wide-area scenes both indoors and outdoors.
The rendering accuracy of novel view synthesis is highly dependent on the accuracy of the camera parameters as well as the performance of the methods themselves.
Even if the rendered RGB images look visually correct, there are many cases where the geometric cues, such as depth maps and normal maps, are inconsistent \cite{ChenLYWXZWLBZ24}.
Therefore, a variety of modalities, including accurate camera parameters and geometric features in addition to RGB images, are essential for the development of novel view synthesis methods that take advantage of omnidirectional images.

\noindent
{\bf 3D Reconstruction} ---
There are many methods for 3D reconstruction from multi-view images, most of which assume that perspective projection images are input.
One of major approach such as Multi-View Stereo (MVS) \cite{SchonbergerZPF16,YaoLLFQ18} is to analyze the similarity between multi-view images, estimate depth maps, and integrate them to reconstruct a 3D model, and another approach is to estimate the distance field \cite{YarivKMGABL20,WangLLTKW21,YarivGKL21,LiMETULL23,ChenLYWXZWLBZ24} and reconstruct a 3D mesh model by applying the marching cubes algorithm \cite{LorensenC87}.
Such methods that assume the input of perspective projection images cannot directly handle omnidirectional images, so it is necessary to introduce a camera model that takes into account the characteristics of the omnidirectional camera.
To address this problem, 360MVSNet \cite{ChiuWSC23} and OmniSDF \cite{KimMJTK24} have been proposed as 3D reconstruction methods from omnidirectional images.
Both methods assume that the omnidirectional camera is approximated by a camera model of a unit sphere, enabling the application of conventional MVS methods.
Research on 3D reconstruction from omnidirectional images is still ongoing, and there are no datasets and standardized benchmarks for performance evaluation.
Currently, researchers build their own datasets and conduct experiments using different evaluation strategies, which makes it difficult to evaluate the accuracy in a uniform manner and to make fair comparisons.
Therefore, it is essential to develop a unified benchmark.

\subsection{Omnidirectional Image Dataset}

There have been several omnidirectional image datasets proposed for various computer vision tasks \cite{ChangDFHNSSZZ17,ArmeniSZS17,HuangCZY22,KimCJK21,CruzHLKBK21,BertelYLR20,WangHCLYSCS19,SongYZCSF17}.
Table \ref{tab:dataset-comparison} shows a summary of existing datasets and a comparison with the OB3D proposed in this paper.
Many datasets are publicly available on the Internet, while some, such as 360Roam \cite{HuangCZY22}, require contacting the authors.
Each dataset provides RGB images, depth maps, normal maps, 3D mesh models, 3D point clouds, etc. In some cases, such as 360Roam \cite{HuangCZY22}, OmniScene \cite{KimCJK21}, and OmniBlender \cite{ChoiK023}, only RGB images are provided.
The image size differs depending on the dataset.
In some cases, such as 360Roam \cite{HuangCZY22}, large images are provided, while in most cases, the image width is from 1,000 to 2,000 pixels.
There are two types of data types: ``Real'' data, which is taken in the real world with a omnidirectional camera, and ``Synthetic'' data, which is based on 3D computer graphics.
Real data faithfully represents a real-world scene, although the camera parameters may contain errors.
On the other hand, synthetic data is suitable for accuracy evaluation since it guarantees ideal camera lenses and accurate camera parameters, although it cannot perfectly reproduce the complex real-world environment.
There are two types of capturing locations: indoor and outdoor, and two types of camera trajectories: Egocentric (Ego.) and Non-Egocentric (Non-ego.).
Egocentric trajectories are circular or spiral trajectories centered on the photographer, and are less burdensome, however, they are not suitable for reconstruction of a wide scene due to the small disparity between the images.
Non-Egocentric trajectories allow free camera movement and wide-area shooting, however, they are more burdensome.
The evaluation criteria vary from each dataset, and evaluation protocols are not provided for many datasets.
To the best of our knowledge, OB3D proposed in this paper is the only dataset that provides an evaluation protocol for 3D reconstruction, and therefore provides an important benchmark for the development of 3D reconstruction methods from omnidirectional images.

\begin{table}[t]
    \caption{Comparison of omnidirectional image datasets, where ``\checkmark'' indicates ``provided'' and ``\xmark'' indicates ``not provided''.}\vspace{-5mm}
    \label{tab:dataset-comparison}
    \begin{center}
    \resizebox{\textwidth}{!}{
    \begin{tabular}{lccccccc}
        \toprule
             Dataset & Matterport3D \cite{ChangDFHNSSZZ17} & Stanford2D-3D-S \cite{ArmeniSZS17} & 360Roam \cite{HuangCZY22} & OmniScene \cite{KimCJK21} & OmniBlender \cite{ChoiK023} & OB3D (Ours) \\
         \cmidrule(rl){1-1}
         \cmidrule(rl){2-7}
             Availability & Public & Public & Limited & Public & Public & Public \\
         \cmidrule(rl){1-1}
         \cmidrule(rl){2-7}
             RGB & \checkmark & \checkmark & \checkmark & \checkmark & \checkmark & \checkmark \\
             Depth & \checkmark & \checkmark & \xmark & \xmark & \xmark & \checkmark \\
             Normal & \checkmark & \checkmark & \xmark & \xmark & \xmark & \checkmark \\
             Mesh/Points & \checkmark / \xmark & \checkmark / \xmark & \xmark / \xmark & \xmark / \xmark & \xmark / \xmark & \xmark / \checkmark \\
         \cmidrule(rl){1-1}
         \cmidrule(rl){2-7}
             Image size [px] & $2,048\times1,024$ & $1,080\times540$ & $6,080\times3,040$ & $1,920\times960$ & $2,000\times1,000$ & $1,600\times800$ \\
             Data source & Real & Real & Real & Real & Synthetic & Synthetic \\
             Environment & Indoor & Indoor & Indoor & Indoor & Indoor/Outdoor & Indoor/Outdoor \\
             Camera trajectory & Non-ego. & Non-ego. & Non-ego. & Ego. & Ego. & Ego./Non-ego.\\
             \# of scenes & 90 & 6 & 10 & 8 & 11 & 12 \\
         \cmidrule(rl){1-1}
         \cmidrule(rl){2-7}
             $^\dagger$Evaluation protocol & 
             \begin{tabular}{c}
                  KM, VOP, SNE \\ RTC, SVL
             \end{tabular}
             & \xmark & \xmark & \xmark & \xmark & 
             \begin{tabular}{c}
                  3D, NVS, CPE
             \end{tabular} \\
        \bottomrule
    \end{tabular}    
    }
    \end{center}
    {\scriptsize
    $\dagger$Evaluation protocol: Keypoint matching (KM), view overlap prediction (VOP), surface normal estimation (SNE), region-type classification (RTC), semantic voxel labeling (SVL), 3D reconstruction (3D), novel view synthesis (NVS), and camera pose estimation (CPE).
    }
\end{table}

\section{Omnidirectional Blender 3D (OB3D) Dataset}

This section describes the overview, dataset creation, and specification of OB3D.

\subsection{Overview}

\begin{figure}[t]
    \centering
    \includegraphics[width=.95\linewidth]{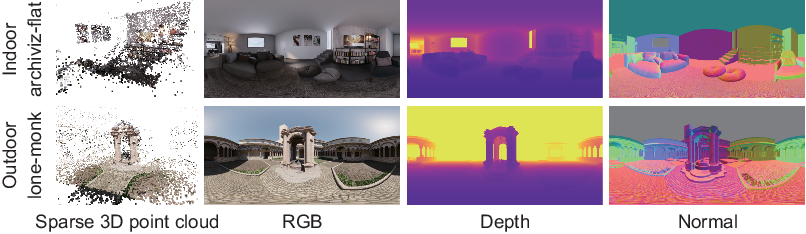}
    \caption{Examples of RGB image, depth map, normal map, and sparse 3D point cloud for indoor and outdoor scenes in OB3D.}
    \label{fig:dataset-examples}
\end{figure}

Omnidirectional Blender 3D (OB3D) is a set of datasets for evaluating 3D reconstruction methods based on 3D scenes created using Blender\footref{fot:blender}.
OB3D was constructed to evaluate the accuracy of mesh models in 3D reconstruction, as well as the rendering quality of RGB images generated by novel view synthesis and the accuracy of camera parameter estimation.
OB3D contains 12 different scenes: {\tt archiviz-flat}, {\tt barbershop}, {\tt bistro}, {\tt classroom}, {\tt emerald-square}, {\tt fisher-hut}, {\tt lone-monk}, {\tt pavillion}, {\tt restroom}, {\tt san-miguel}, {\tt sponza}, and {\tt sun-temple}.
OB3D covers a variety of indoor and outdoor environments; 5 of the 12 scenes are indoor scenes and 7 are outdoor scenes.
Since an omnidirectional camera can capture images of its own entire surroundings, the target area is a large space.
Therefore, it is important for OB3D to include many indoor and outdoor scenes to expand the versatility and range of application of the 3D reconstruction methods.
The data for each scene consists of omnidirectional images rendered with ideal equirectangular projection without lens distortion, corresponding accurate camera parameters, depth maps, normal maps, and sparse 3D point clouds, where the sparse 3D point cloud was obtained using OpenMVG\footref{fot:openmvg}.
Fig. \ref{fig:dataset-examples} shows RGB images, depth maps, normal maps, and sparse 3D point clouds for some scenes in OB3D.

\begin{figure}[t]
    \centering
    \includegraphics[width=.95\linewidth]{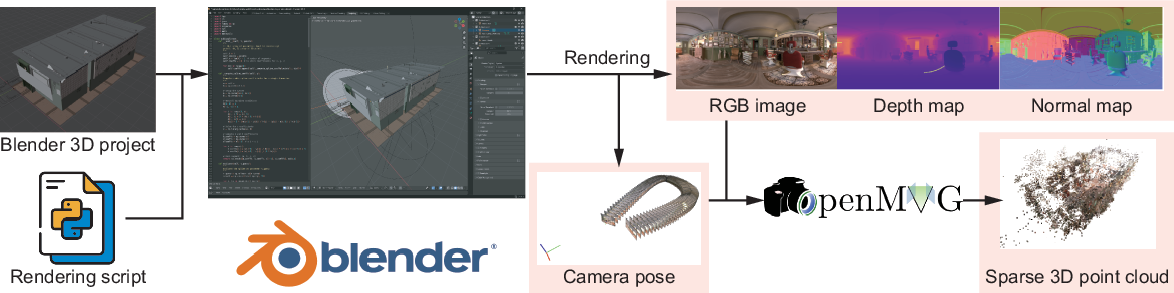}
    \caption{Pipeline of creating OB3D: RGB images (omnidirectional images), depth maps, normal maps, camera parameters for each image, and sparse 3D point cloud obtained using OpenMVG from a Bldender 3D project.}
    \label{fig:data-generation-pipeline}
\end{figure}

\subsection{Data Creation}

This section describes the procedure for creating OB3D.
Fig. \ref{fig:data-generation-pipeline} shows the pipeline for creating OB3D datasets.
First, the Blender 3D Project and a script for rendering using the Blender Python API\footnote{\url{https://docs.blender.org/api/current/}\label{fot:blenderpython}} are loaded in Blender. 
We use the 12 Blender 3D projects in creating OB3D.
The details of rendering using the Blender Python API are discussed in Sect. \ref{sec:rendering}.
Next, while moving the camera position in Blender, the RGB images, depth maps, and normal maps of the omnidirectional images are rendered, and the camera parameters for each view are output.
Finally, using the camera parameters and the RGB images, OpenMVG\footref{fot:openmvg} performs SfM to obtain a sparse 3D point cloud.
For detailed information about the OpenMVG procedure, please see the supplementary material.
Unlike OpenSfM\footref{fot:opensfm}, OpenMVG can perform SfM using pre-defined camera parameters.
Therefore, it is possible to execute SfM using the exact camera parameters output from Blender.
360Roam \cite{HuangCZY22} and ODGS \cite{LeeCHL24} also provide sparse 3D point clouds obtained using OpenMVG.
For example, a sparse 3D point cloud can be used for novel view synthesis using 3DGS.

\subsection{Rendering Using Blender Python API}
\label{sec:rendering}

\begin{figure}[t]
    \centering
    \includegraphics[width=.95\linewidth]{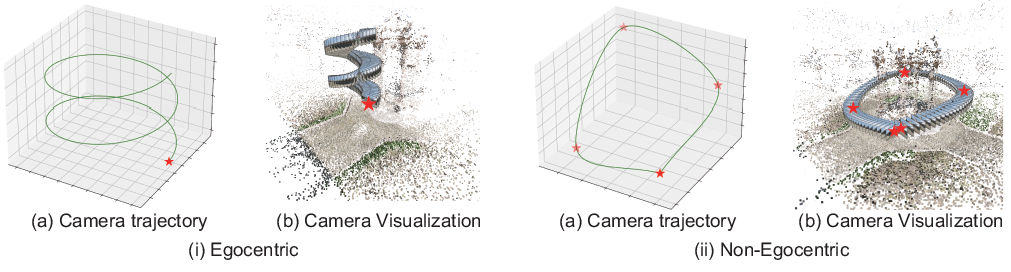}
    \caption{Camera trajectories in OB3D: (i) Egocentric trajectory and (ii) Non-Egocentric trajectory.}
    \label{fig:camera-settings}
\end{figure}

Blender can define a camera object in 3D space and render an image as observed from that camera.
The viewing position from which the image is rendered can be set to (i) an Egocentric trajectory or (ii) a Non-Egocentric trajectory, as shown in Fig. \ref{fig:camera-settings}.
The Egocentric trajectory is to capture omnidirectional images with the photographer at the center, as if the camera were rotating around the photographer.
Omnidirectional images taken with a camera trajectory spiraling around a certain point are also considered to be an Egocentric trajectory.
Most of the 3D reconstruction methods and novel view synthesis methods that have been proposed using omnidirectional images assume that the input images are captured using the Egocentric trajectory.
The Non-Egocentric trajectory is to capture omnidirectional images by moving freely within a scene.
When capturing omnidirectional images with Egocentric/Non-Egocentric trajectories in the real world, the camera orientation changes depending on the camera position, and since the photographer holds the camera in his/her hand, the camera orientation rarely changes up and down.
If $N$ images are to be rendered for each camera trajectory, $N+1$ camera positions are set on the trajectory.
The reason for setting $N+1$ camera positions is to find the current camera position based on the next camera position.
The camera orientation at a camera position $i\in \{1,\cdots,N\}$ is defined as the direction vector to the next camera position $i+1$ projected onto the $X$-$Y$ plane of the world coordinate system.
Using the camera positions and orientations obtained above, RGB images, depth maps and normal maps are rendered as omnidirectional images.
Please refer to the supplemental material for more details on rendering and camera orientation definition.

\subsection{Specification and Statistics}

OB3D is a dataset for evaluating 3D reconstruction that includes various modalities other than omnidirectional images.
Table \ref{tab:data-statistics} shows the specifications and statistics of OB3D.
OB3D covers a variety of indoor and outdoor environments and consists of 12 different scenes.
Each scene contains a subset of images rendered with Egocentric trajectories and Non-Egocentric trajectories, respectively.
Each camera trajectory contains 100 multi-view omnidirectional images, and each image size is $1,600 \times 800$ pixels.
RGB images, depth maps, normal maps, sparse 3D point clouds, and camera parameters are provided for each viewpoint.
The RGB images are provided in PNG format, and the depth and normal images are provided in OpenEXR format.
A sparse 3D point cloud reconstructed using OpenMVG is provided both for Egocentric and Non-Egocentric trajectories.
The number of reconstructed points varies from scene to scene. However, thousands to tens of thousands of points are provided and can be used to initialize Gaussians in 3DGS.
To evaluate the tasks of camera parameter estimation, novel view synthesis, and 3D reconstruction, OB3D uses 25 out of 100 images from each scene for training and 25 for evaluation, as in OmniBlender \cite{ChoiK023}.
The training data are the 25 images with an image index of $[0, 4, 8, \cdots, 96]$, and the evaluation data are the 25 images with an image index of $[2, 6, 10, \cdots, 98]$.

\begin{table}[t]
    \caption{The statistics and data details of OB3D dataset. 'I' and 'O' indicate Indoor and Outdoor, respectively.}
    \label{tab:data-statistics}
    \centering
    \small
    \resizebox{\textwidth}{!}{
        \begin{tabular}{ccccccccccc}
            \toprule
                &  &  &  & \multicolumn{3}{c}{Format} & Split & \multicolumn{2}{c}{\# of 3D points} \\
            \cmidrule(rl){5-7}
            \cmidrule(rl){8-8}
            \cmidrule(rl){9-10}
               Scene & Environment & \# of images & Image size [px] & RGB & Depth & Normal & train/test & Egocentric & Non-Egocentric \\
            \cmidrule(r){1-1}
            \cmidrule(rl){2-2}
            \cmidrule(rl){3-8}
            \cmidrule(l){9-10}
               {\tt archiviz-flat} & I & \multirow{12}{*}{100} & \multirow{12}{*}{1,600$\times$800} & \multirow{12}{*}{PNG} & \multirow{12}{*}{OpenEXR} & \multirow{12}{*}{OpenEXR} & \multirow{12}{*}{25/25} & 13,216 & 9,537 \\
               {\tt barbershop} & I & & & & & & & 43,164 & 40,635 \\
               {\tt bistro} & O & & & & & & & 48,554 & 79,242 \\
               {\tt classroom} & I & & & & & & & 21,802 & 23,343 \\
               {\tt emerald-square} & O & & & & & & & 34,835 & 23,982 \\
               {\tt fisher-hut} & O & & & & & & & 6,087 & 3,859 \\
               {\tt lone-monk} & O & & & & & & & 55,733 & 61,013 \\
               {\tt pavillion} & I & & & & & & & 31,686 & 43,855 \\
               {\tt restroom} & I & & & & & & & 23,754 & 31,992 \\
               {\tt san-miguel} & O & & & & & & & 52,582 & 47,088 \\
               {\tt sponza} & O & & & & & & & 53,340 & 65,539 \\
               {\tt sun-temple} & O & & & & & & & 28,322 & 37,932\\
            \bottomrule
        \end{tabular}
    }
\end{table}

\subsection{Evaluation protocol}
\label{sec:eval}

OB3D provides three evaluation protocols: (i) camera parameter estimation, (ii) novel viewpoint synthesis, and (iii) 3D reconstruction.
An overview of each evaluation protocol is given below.

\noindent
{\bf (i) Camera Parameter Estimation} ---
In the evaluation of camera parameter estimation, the camera parameters are estimated using the multi-view omnidirectional image of OB3D, and the estimation error is evaluated against the ground-truth camera parameters provided in OB3D.
Since the estimated camera parameters differ in scale and position from those in OB3D, a direct comparison is not meaningful.
Therefore, we evaluate the estimation error using the relative camera parameters between an arbitrary pair of views instead of the error for each individual view.
The relative camera parameters are evaluated using the Relative Rotation Error (RRA) and Relative Translation Error (RTA), which measure the angular error between the rotation matrix and the translation vector, respectively.
We also employ AUC@5, which indicates the percentage of all image pairs in which both RRA and RTA are less than 5 degrees, as an evaluation metric.
The displacement of the trajectory can be evaluated by aligning the scale and position of the estimated camera trajectory with those of the OB3D camera trajectory.
To evaluate the accuracy of the camera trajectory, the scale and position of the two camera trajectories are aligned, and the Absolute Trajectory Error (ATE), which calculates the Root Mean Squared Error (RMSE) between them, is used as an evaluation metric.

\noindent
{\bf (ii) Novel View Synthesis} ---
In the evaluation of the novel view synthesis, the radiance field of a scene is estimated using RGB images, camera parameters, and a sparse 3D point cloud, and the quality of the rendered images is evaluated.
The 25 images included in the training data and the 25 images included in the test data for each scene are used in the evaluation to verify the generalization performance of novel view synthesis models.
According to the evaluation criteria of the novel view synthesis methods using omnidirectional images \cite{ChoiK023,LiHYC24}, the quality of the rendered images is evaluated using PSNR, SSIM, LPIPS (A), and LPIPS (V) in RGB space, where LPIPS (A) and LPIPS (V) indicate that AlexNet \cite{KrizhevskySH12} and VGG \cite{LiuD15} are used to extract image features, respectively.

\noindent
{\bf (iii) 3D Reconstruction} ---
In the evaluation of 3D reconstruction, the camera parameters and the omnidirectional images are used to reconstruct a 3D mesh model, and the errors between the rendered depth map and the depth map provided by OB3D are measured.
Although measuring the distance between mesh models is a standard approach for evaluating 3D reconstruction, a direct comparison between mesh models cannot provide a valid evaluation since Blender 3D Project has mesh models in regions not included in the rendering of omnidirectional images. 
Therefore, depth maps are rendered from the reconstructed mesh models, and the accuracy of the 3D reconstruction is evaluated by the error between the depth maps.
In addition to Root Mean Squared Error (RMSE) and Mean Absolute Error (MAE), Absolute Relative Error (AbsRel) and $\delta_{1.25}$, which are widely used to evaluate the accuracy of depth map estimation, are used as evaluation metrics. 
For fair evaluation, regions with extremely large depth values, such as ``sky'' regions, are excluded from the evaluation target.

\section{Experiments}
\label{sec:experiments}

In this section, we conduct experiments using OB3D to evaluate the accuracy of methods for tasks that use omnidirectional images as input, such as camera parameter estimation, novel viewpoint synthesis, and 3D reconstruction, and demonstrate that OB3D can provide a benchmark for topics that make use of omnidirectional images. 
For the details of experimental results, refer to the supplementary material.
Camera parameter estimation, novel view synthesis, and part of 3D reconstruction are conducted on NVIDIA GeForce RTX 4090 (24GB), while the rest of 3D reconstruction is conducted on NVIDIA A100 (80GB) PCIe.

\subsection{Camera Parameter Estimation}
\label{sec:exp_cpe}

In this experiment, we use OpenSfM\footref{fot:opensfm} and OpenMVG\footref{fot:openmvg} used in SfM for omnidirectional images as a baseline.
By setting the camera model as an omnidirectional camera when loading images, the camera parameters can be estimated from omnidirectional images.
Table \ref{tab:quantitabe-result} shows the results of evaluating the accuracy of camera parameter estimation for all scenes in OB3D.
For indoor scenes, both methods can estimate camera parameters that are close to those provided by OB3D, regardless of the camera trajectories.
For outdoor scenes, both methods can estimate the camera parameters with high accuracy for Egocentric trajectories, while the estimation accuracy decreases for Non-Egocentric trajectories.
In particular, the estimation accuracy is significantly degraded for Non-Egocentric trajectories in scenes containing symmetrical structures such as {\tt sponza}.
For more detailed results of the camera parameter estimation for each scene, refer to the supplementary material.

\begin{table}[t]
    \centering
    \caption{Experimental results of camera parameter estimation, where each value indicates the average of all scenes in OB3D.}
    \label{tab:quantitabe-result}
    \resizebox{\textwidth}{!}{
        \begin{tabular}{cccccccccc}
            \toprule
             &  & \multicolumn{4}{c}{Egocentric} & \multicolumn{4}{c}{Non-Egocentirc} \\
             Scene & Method & RRA$\downarrow$ & RTA$\downarrow$ & AUC@5$\uparrow$ & ATE$\downarrow$ & RRA$\downarrow$ & RTA$\downarrow$ & AUC@5$\uparrow$ & ATE$\downarrow$ \\
            \cmidrule(rl){1-1}
            \cmidrule(rl){2-2}
            \cmidrule(rl){3-6}
            \cmidrule(rl){7-10}
            \multirow{2}{*}{Indoor}  & OpenSfM\footref{fot:opensfm} & {\bf 0.405} & 0.037 & {\bf 1.000} & {\bf 0.000} & {\bf 0.405} & {\bf 0.022} & {\bf 1.000} & {\bf 0.000} \\
                                     & OpenMVG\footref{fot:openmvg} & {\bf 0.405} & 0.065 & {\bf 1.000} & {\bf 0.000} & 0.413 & 0.058 & {\bf 1.000} & {\bf 0.000} \\
            \cmidrule(rl){1-1}
            \cmidrule(rl){2-2}
            \cmidrule(rl){3-6}
            \cmidrule(rl){7-10}
            \multirow{2}{*}{Outdoor} & OpenSfM\footref{fot:opensfm} & {\bf 0.405} & {\bf 0.033} & {\bf 1.000} & {\bf 0.000} & 0.406 & 0.033 & {\bf 1.000} & {\bf 0.000} \\
                                     & OpenMVG\footref{fot:openmvg} & {\bf 0.405} & 0.077 & {\bf 1.000} & {\bf 0.000} & 12.386 & 6.443 & 0.851 & 0.091 \\    
            \bottomrule
        \end{tabular}
    }
\end{table}

\subsection{Novel View Synthesis}
\label{sec:exp_nvs}

In this experiment, we use EgoNeRF \cite{ChoiK023} based on NeRF \cite{MildenhallSTBRN20} and ODGS \cite{LeeCHL24}, op43dgs \cite{HuangBGLG24}, and OmniGS \cite{LiHYC24} based on 3DGS \cite{KerblKLD23} as the baselines for novel view synthesis.
EgoNeRF generates a novel view image from an omnidirectional image taken with an Egocentric camera trajectory.
On the other hand, 3DGS-based methods are designed to rasterize 3D Gaussians on the unit sphere and can take advantage of images from Egocentric and Non-Egocentric camera trajectories.
In addition to RGB images and camera parameters, these methods use sparse 3D point clouds as initialization for radiance field estimation.
Table \ref{tab:nvs-quantitative-result} shows the experimental results for each method.
EgoNeRF, which takes Egocentric images as input, demonstrates high rendering accuracy, and robustness to environmental changes.
On the other hand, the 3DGS-based method varies its rendering accuracy by 4.8--30.7\% when the environment changes from indoor to outdoor, and by 2.4--24.2\% when the camera trajectory changes from Egocentric to Non-Egocentric.
In particular, op43dgs shows the same level of accuracy as the other methods for outdoor Non-Egocentric images, however, the variation of accuracy is large for changes in environment and trajectory.
From the above, the use of OB3D enables a detailed analysis of the performance of each method by taking into account the diversity of scene environments and camera trajectories.

\begin{table}[t]
    \centering
    \caption{Experimental results of novel view synthesis.
    The values indicate the average of each metric.
    EgoNeRF is not applicable to Non-Egocentric images because it assumes Egocentric image inputs.}
    \label{tab:nvs-quantitative-result}
    \resizebox{\textwidth}{!}{
    \begin{tabular}{cccccccccccccc}
         \toprule
             & \multicolumn{6}{c}{Indoor} & \multicolumn{6}{c}{Outdoor} \\
         \cmidrule(rl){2-7}
         \cmidrule(rl){8-13}
             & \multicolumn{3}{c}{Egocentric} & \multicolumn{3}{c}{Non-Egocentric} & \multicolumn{3}{c}{Egocentric} & \multicolumn{3}{c}{Non-Egocentric} \\
         \cmidrule(rl){2-4}
         \cmidrule(rl){5-7}
         \cmidrule(rl){8-10}
         \cmidrule(rl){11-13}
             Method & PSNR [dB] & SSIM & LPIPS (A) & PSNR [dB] & SSIM & LPIPS (A) & PSNR [dB] & SSIM & LPIPS (A) & PSNR [dB] & SSIM & LPIPS (A) \\
         \cmidrule(rl){1-1}
         \cmidrule(l){2-13}
            EgoNeRF \cite{ChoiK023} 
            & 32.26 & {\bf 0.906} & {\bf 0.128} & --- & --- & --- & {\bf 32.18} & {\bf 0.916} & {\bf 0.086} & --- & --- & --- \\
            ODGS \cite{LeeCHL24} 
            & 28.14 & 0.840 & 0.241 & 26.80 & 0.819 & 0.256 & 27.47 & 0.849 & 0.176 & 25.04 & 0.771 & {\bf 0.243} \\
            op43dgs \cite{HuangBGLG24} 
            & 23.86 & 0.780 & 0.402 & {\bf 31.18} & {\bf 0.905} & {\bf 0.154} & 19.22 & 0.685 & 0.407 & 25.11 & {\bf 0.828} & 0.244 \\
            OmniGS \cite{LiHYC24} 
            & {\bf 33.25} & 0.897 & 0.169 & 27.93 & 0.839 & 0.247 & 29.89 & 0.907 & 0.112 & {\bf 25.86} & 0.785 & 0.258 \\
         \bottomrule
    \end{tabular}
    }
\end{table}

\subsection{3D Reconstruction}
\label{sec:exp_3dr}

We evaluate the accuracy of 3D reconstruction of COLMAP \cite{SchonbergerZPF16}, NeuS \cite{WangLLTKW21}, and OmniSDF \cite{KimMJTK24} using the camera parameters and omnidirectional images included in OB3D.
Note that the details of the experimental condition and results of OmniSDF are available in the supplementary material.
COLMAP is an MVS method that assumes perspective projection images as input and cannot directly process omnidirectional images, so they are converted to cube maps and then reconstructed in 3D.
NeuS is a method for 3D reconstruction from multi-view images that combines implicit function representation and radiance field estimation, however, it cannot handle omnidirectional images directly, so the data loader in the official implementation is modified to handle them.
NeuS uses the marching cubes algorithm \cite{LorensenC87} to reconstruct the mesh after training is complete, where the voxel space size is $1024^3$ in the experiment.
As in Sect. \ref{sec:eval}, we evaluate the accuracy of the 3D reconstruction with a depth map.
Table \ref{tab:quantitative-3d-recontruction} shows the experimental results for 3D reconstruction.
The accuracy of 3D reconstruction is lower for outdoor scenes than for indoor scenes, and is easily affected by noise in a wide-area space.
When we focus on the effect of the difference in camera trajectories, the accuracy is slightly higher in the case of a Non-Egocentric trajectory.
In addition, the accuracy is significantly lower for scenes with a large reconstruction area, such as {\tt emerald-square} and {\tt fisher-hut} scenes.
From the above, the evaluation using OB3D revealed that the size of the scene and the camera trajectory have a significant effect on the accuracy of the 3D reconstruction.

\begin{table}[t]
    \centering
    \caption{Experimental results of 3D Reconstruction. NeuS* means a modification to NeuS \cite{WangLLTKW21} for using omnidirectional images as input. 
    Results for OmniSDF are available in the supp. material.}
    \label{tab:quantitative-3d-recontruction}
    \resizebox{\textwidth}{!}{
    \begin{tabular}{cccccccccc}
         \toprule
             & & \multicolumn{4}{c}{COLMAP \cite{SchonbergerZPF16}} & \multicolumn{4}{c}{NeuS*} \\
         \cmidrule(rl){3-6}
         \cmidrule(rl){7-10}
             & & \multicolumn{2}{c}{Egocentric} & \multicolumn{2}{c}{Non-Egocentric} & \multicolumn{2}{c}{Egocentric} & \multicolumn{2}{c}{Non-Egocentric} \\
         \cmidrule(rl){1-2}
         \cmidrule(rl){3-4}
         \cmidrule(rl){5-6}
         \cmidrule(rl){7-8}
         \cmidrule(rl){9-10}
             Type & Scene & RMSE [m] $\downarrow$ & $\delta_{1.25}$ [\%] $\uparrow$ & RMSE [m] $\downarrow$ & $\delta_{1.25}$ [\%] $\uparrow$ & RMSE [m] $\downarrow$ & $\delta_{1.25}$ [\%] $\uparrow$ & RMSE [m] $\downarrow$ & $\delta_{1.25}$ [\%] $\uparrow$ \\
         \cmidrule(rl){1-2}
         \cmidrule(l){3-10}
             \multirow{6}{*}{\rotatebox{90}{Indoor}}
             & {\tt archiviz-flat}  & 0.883 & 0.688 & 1.073 & 0.635 & 0.206 & {\bf 0.994} & 0.191 & {\bf 0.986} \\
             & {\tt barbershop}     & {\bf 0.277} & 0.960 & {\bf 0.240} & 0.950 & 0.319 & 0.991 & {\bf 0.121} & {\bf 0.986} \\
             & {\tt classroom}      & 0.520 & 0.953 & 0.725 & 0.855 & {\bf 0.175} & 0.991 & 0.189 & 0.982 \\
             & {\tt restroom}       & 2.359 & 0.936 & 1.152 & 0.969 & 0.423 & 0.989 & 0.499 & 0.985 \\
             & {\tt sun-temple}     & 0.848 & {\bf 0.987} & 0.776 & {\bf 0.986} & 0.908 & 0.978 & 0.797 & 0.985 \\
             & Average (Indoor)     & 0.978 & 0.905 & 0.793 & 0.879 & 0.406 & 0.989 & 0.359 & 0.983\\
         \cmidrule(rl){1-2}
         \cmidrule(l){3-10}
             \multirow{8}{*}{\rotatebox{90}{Outdoor}}
             & {\tt bistro}         & 1.842 & {\bf 0.981} & 1.802 & 0.972 & 2.070 & {\bf 0.970} & 2.348 & 0.970 \\
             & {\tt emerald-square} & 3.738 & 0.979 & 3.869 & {\bf 0.975} & 6.235 & 0.951 & 3.772 & {\bf 0.976} \\
             & {\tt fisher-hut}     & 3.068 & 0.931 & 2.779 & 0.885 & 2.975 & 0.938 & 3.345 & 0.949 \\
             & {\tt lone-monk}      & 1.664 & 0.928 & 1.316 & 0.912 & 1.251 & 0.912 & 1.322 & 0.925 \\
             & {\tt pavillion}      & 2.281 & 0.963 & 2.515 & 0.967 & 2.522 & 0.952 & 2.518 & 0.958 \\
             & {\tt san-miguel}     & 2.009 & 0.835 & 2.327 & 0.781 & 1.357 & 0.826 & 1.187 & 0.872 \\
             & {\tt sponza}         & {\bf 1.065} & 0.933 & {\bf 0.984} & 0.922 & {\bf 1.311} & 0.897 & {\bf 1.148} & 0.892 \\
             & Average (Outdoor)    & 2.238 & 0.936 & 2.228 & 0.916 & 2.532 & 0.921 & 2.234 & 0.934 \\
         \cmidrule(rl){1-2}
         \cmidrule(l){3-10}
             & Average (all)        & 1.608 & 0.920 & 1.510 & 0.898 & 1.469 & 0.955 & 1.297 & 0.959 \\
         \bottomrule
    \end{tabular}
    }
\end{table}

\section{Limitations}
\label{sec:limitation}

OB3D is a set of synthetic datasets consisting of omnidirectional images rendered using Blender, and is suitable for research and performance evaluation of methods since it is free from camera lens distortion and provides accurate camera parameters.
On the other hand, it does not perfectly represent real-world data since it differs from the actual environment and camera acquisition.
In the future, we plan to expand the diversity of scenes based on OB3D by adding geometric information such as RGB images, depth maps, and normals taken in the real world.
Since OB3D scenes cover a wide area, direct comparison with 3D mesh models rendered by Blender is difficult, and currently only the area that appears in the omnidirectional images was used for evaluation.
If the reconstructed mesh model can be compared with the ground truth, a more accurate evaluation will be possible.

\section{Conclusion}

In this paper, we proposed the Omnidirectional Blender 3D (OB3D) dataset as a dataset and benchmark for research and evaluation of 3D reconstruction using omnidirectional images.
OB3D contains omnidirectional RGB images generated by Blender, as well as accurate camera parameters, depth maps, and normal maps, and can be used as a standard evaluation platform not only for 3D reconstruction but also for camera parameter estimation and novel view synthesis.
In the future, we plan to integrate the real-world RGB images into OB3D to expand the data diversity.

\section{Acknowledgment}

All data used in OB3D were created based on Blender 3D projects that have been licensed for non-commercial use.
We would like to thank the creators of each project.
We also thank Kent Selwyn The for reviewing OB3D and the evaluation codes.
This work was supported in part by JSPS KAKENHI JP 23H00463 and 25K03131, and JST BOOST, Japan Grant Number JPMJBS2421.

\clearpage
\appendix

\section{OpenMVG Procedure}

The procedure for reconstructing a sparse 3D point cloud using OpenMVG is as follows.
\begin{enumerate}
[leftmargin=5mm]
\item The input image is registered and the camera is configured.
Since the camera parameters of OB3D are different from those of OpenMVG, it is necessary to perform the appropriate conversion and write them into a file format that OpenMVG can handle.
\item Feature points are extracted and feature point matching between images is performed.
To increase the number of reconstructed point clouds, the hyperparameters should be adjusted for optimization.
Refer to Sect. \ref{sec:cpe_set} for setting details.
\item A sparse 3D point cloud is reconstructed and camera parameters are estimated.
Since the resulting point cloud includes points indicating the camera position, a process to erase the camera is necessary to extract only the scene information.
\end{enumerate}
For the concrete execution commands and processing details, please refer to the code available on GitHub\footref{fot:github}.
This paper employs OpenMVG version 2.1.0\footref{fot:openmvg}.

\section{Rendering Using Blender Python API}

OB3D is created by automatically rendering omnidirectional RGB images, depth maps, and normal maps from the Blender 3D Project using the Python API.
An arbitrary number of cameras are sampled from Egocentric or Non-Egocentric camera trajectories, and rendering is performed from each view.
For details of the camera trajectory calculation, refer to Sect. \ref{sec:camera_traj}.
The camera coordinate system used in general computer vision is different from the camera coordinate system in Blender in terms of the coordinate directions.
In OpenCV, a major computer vision library, $+X$ faces right, $+Y$ faces down, and $+Z$ faces forward, while in Blender, $+X$ faces right, $+Y$ faces up, and $+Z$ faces backward.
Therefore, post-processing is required when converting camera convention from Blender to OpenCV.
All extrinsic parameters of the camera in OB3D represent the conversion from the world coordinate system to the camera coordinate system.
The camera parameters are stored in JSON, the RGB images in PNG, and the depth maps and normal maps in OpenEXR.
For the detailed implementation of rendering script using Blender Python API, refer to the file {\tt generate\_trajectory-data.py} in the GitHub repository\footref{fot:github}.

\section{Details of OB3D}

We describe the details of OB3D in this section.

\subsection{License}

Table \ref{tab:license} shows the licenses of the Blender 3D Project corresponding to each scene.
OB3D is released as CC BY-NC-SA 4.0 to include all data licenses.
The URLs where the Blender 3D Project for each scene can be downloaded are as follows.
\begin{itemize}
[leftmargin=5mm]
    \small
    \item {\tt archiviz-flat}\\
    \url{https://download.blender.org/demo/cycles/flat-archiviz.blend}
    \item {\tt barbershop}\\
    \url{https://svn.blender.org/svnroot/bf-blender/trunk/lib/benchmarks/cycles/barbershop_interior/}
    \item {\tt bistro}\\
    \url{https://developer.nvidia.com/bistro}
    \item {\tt classroom}\\
    \url{https://download.blender.org/demo/test/classroom.zip}
    \item {\tt emerald-square}\\
    \url{https://developer.nvidia.com/emerald-square}
    \item {\tt fisher-hut}\\
    \url{https://www.blendswap.com/blend/30099}
    \item {\tt lone-monk}\\
    \url{https://download.blender.org/demo/cycles/lone-monk_cycles_and_exposure-node_demo.blend}
    \item {\tt pavillion}\\
    \url{https://download.blender.org/demo/test/pabellon_barcelona_v1.scene_.zip}
    \item {\tt restroom}\\
    \url{https://blendswap.com/blend/14216}
    \item {\tt san-miguel}\\
    \url{https://casual-effects.com/g3d/data10/research/model/San_Miguel/San_Miguel.zip}
    \item {\tt sponza}\\
    \url{https://casual-effects.com/g3d/data10/common/model/crytek_sponza/sponza.zip}
    \item {\tt sun-temple}\\
    \url{https://developer.nvidia.com/sun-temple}
\end{itemize}

\begin{table}[t]
    \caption{License and data source for each scene included in OB3D.}
    \label{tab:license}
    \centering
    \begin{tabular}{lll}
        \toprule
            Scene & License & Data Source\\
        \cmidrule{1-3}
            {\tt archiviz-flat}  & CC-BY 4.0 & Blender Demo\\
            {\tt barbershop}     & CC-BY 4.0 & Blender Demo\\
            {\tt bistro} [a]        & CC-BY 4.0 & NVIDIA Developer\\
            {\tt classroom}      & CC0 1.0 & Blender Demo\\
            {\tt emerald-square} [b] & CC BY-NC-SA 3.0 & NVIDIA Developer\\
            {\tt fisher-hut}     & CC-BY 4.0 & Blend Swap\\
            {\tt lone-monk}      & CC-BY 4.0 & Blender Demo\\
            {\tt pavillion}      & CC-BY 4.0 & Blender Demo\\
            {\tt restroom}       & CC-BY 3.0 & Blend Swap\\
            {\tt san-miguel}     & CC-BY 3.0 & McGuire\\
            {\tt sponza}         & CC-BY 3.0 & McGuire\\
            {\tt sun-temple} [c]     & CC BY-NC-SA 4.0 & NVIDIA Developer\\
        \bottomrule
    \end{tabular}
\end{table}

\subsection{Data in OB3D}

\begin{figure}
    \centering
    \includegraphics[width=\linewidth]{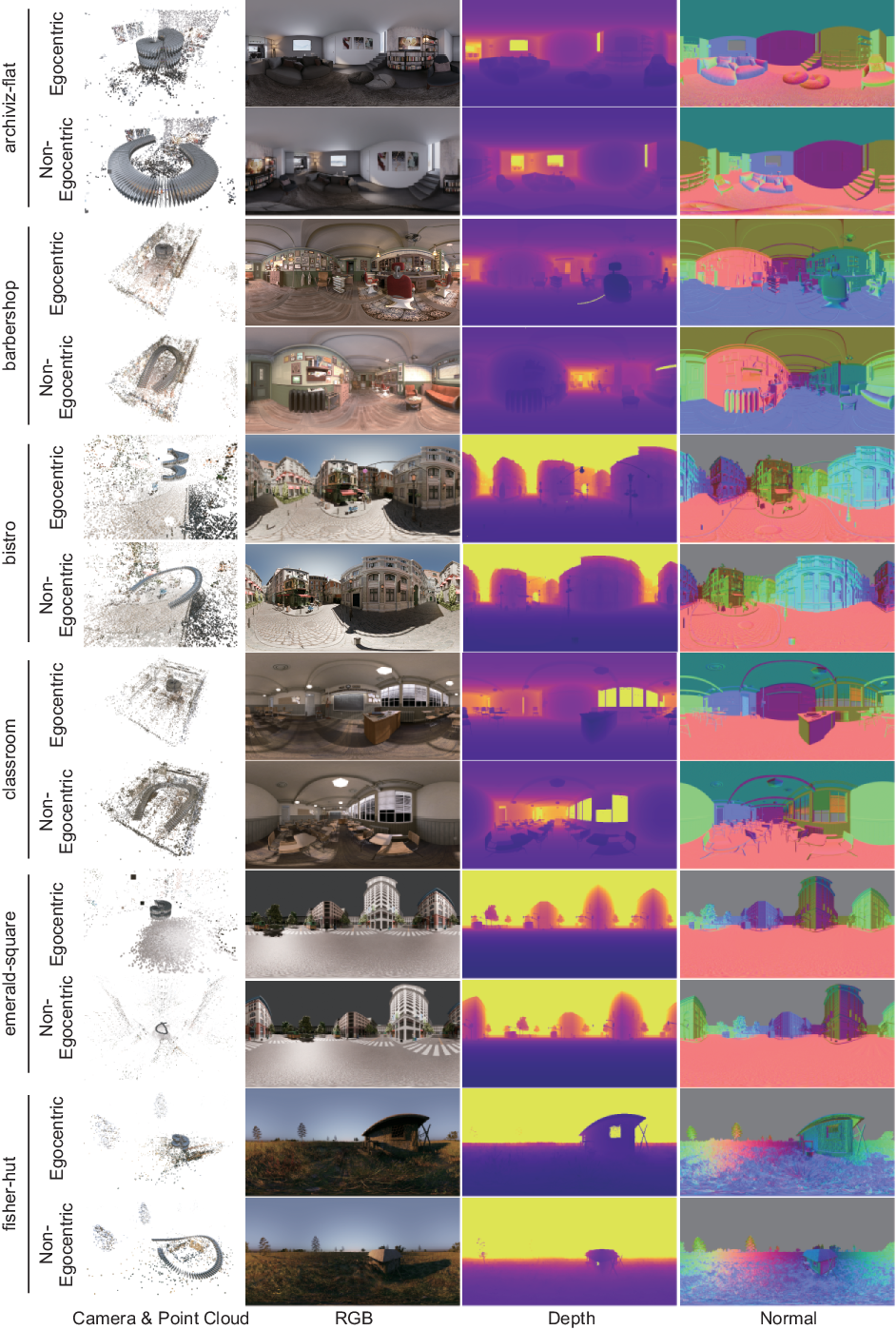}
    \caption{Data included in each scene of OB3D.}
    \label{fig:supp-all-scenes-part1}
\end{figure}

\begin{figure}
    \centering
    \includegraphics[width=\linewidth]{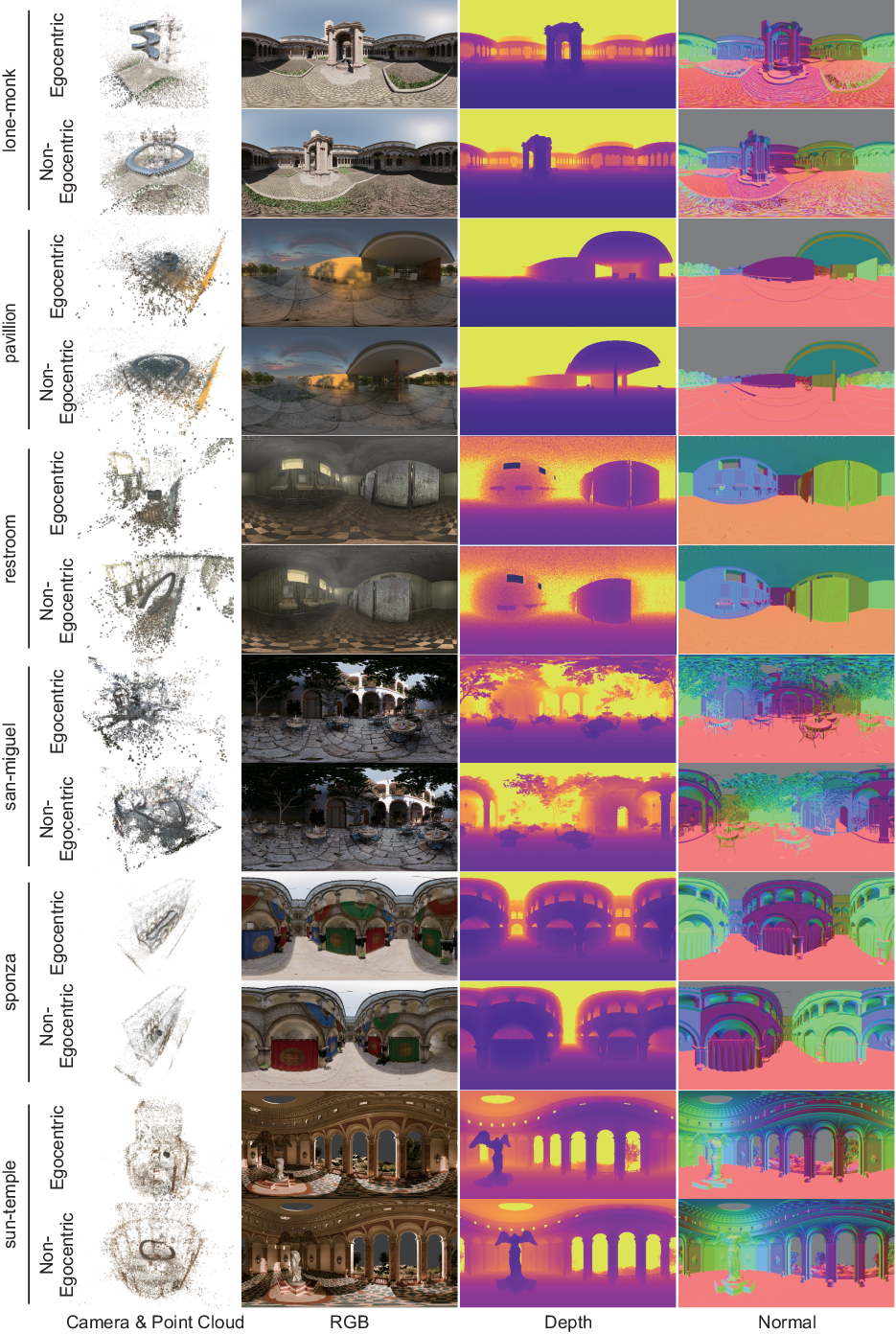}
    \caption{Data included in each scene of OB3D (Continued).}
    \label{fig:supp-all-scenes-part2}
\end{figure}

OB3D consists of 12 indoor/outdoor scenes as shown in Figs. \ref{fig:supp-all-scenes-part1} and \ref{fig:supp-all-scenes-part2}.
Each scene contains data rendered by two camera trajectories, Egocentric and Non-Egocentric, where camera parameters and sparse 3D point clouds are reconstructed by OpenMVG, as well as omnidirectional RGB images, depth maps, and normal maps.


\subsection{Egocentric/Non-Egocentric Camera Trajectory}
\label{sec:camera_traj}

We describe the Egocentric and Non-Egocentric trajectories of the omnidirectional camera used in OB3D.
For the detailed implementation, please refer to the file {\tt generate\_trajectory-data.py} in the GitHub repository\footref{fot:github}.

In the Egocentric trajectory, the start position of an arbitrary camera trajectory is first decided in the Blender 3D Project.
Next, a circular trajectory is taken from the start position, parallel to the $X$-$Y$ plane and centered at a point $(-r/2, 0, 0)$ away from the start position at a specified distance of $r$.
The number of circular trajectories and the amount of ascent along the $Z$ axis at each trajectory are specified to form an Egocentric trajectory.
By sampling an arbitrary number of camera positions on the Egocentric trajectory, an arbitrary number of camera positions can be created.

In the Non-Egocentric trajectory, an arbitrary number of points (keypoints) that should be passed in the creation of camera trajectories are first set and the order of passing is decided.
Next, a smooth trajectory that passes through each keypoint in the specified order is calculated using cubic spline interpolation.
Finally, the camera position can be flexibly set by sampling an arbitrary number of camera positions on the Non-Egocentric trajectory.

\subsection{Rotation Matrix}

\begin{figure}
    \centering
    \includegraphics[width=\linewidth]{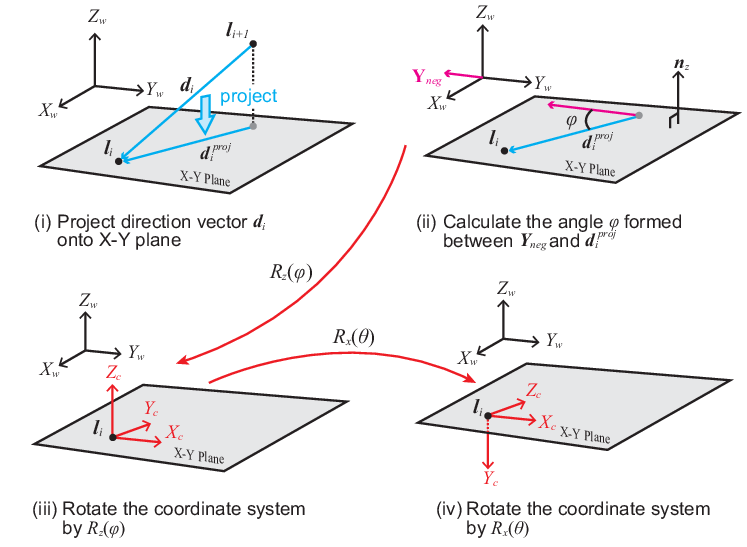}
    \caption{Calculation for the rotation matrix $\bm{R}_i$ of a camera $C_i$.}
    \label{fig:camera-orientation}
\end{figure}

Fig. \ref{fig:camera-orientation} shows an overview of the calculation of the rotation matrix $\bm{R}_i$ of a camera on a camera trajectory.
When $N$ cameras are defined on a camera trajectory, they are determined by sampling the 3D positions of $N+1$ points on the trajectory.
The position of the $i$-th camera $C_i$ on the camera trajectory is given by the $i$-th 3D position $\bm{l}_i$.
Also, let $\bm{d}_i = \bm{l}_i - \bm{l}_{i+1}$ be the vector from the $i+1$-th sample point to the $i$-th sample point.
When using an omnidirectional camera to capture a video image, the camera is often held parallel to the ground.
Therefore, OB3D assumes that the camera orientation is always horizontal to the $X$-$Y$ plane of the world coordinate system.
The vector $\bm{d}^{proj}_i$ of the projection of $\bm{d}_i$ onto the $X$-$Y$ plane is calculated by
\begin{equation}
    \bm{d}^{proj}_i = \bm{d}_i - \bm{n}_z(\bm{n}_z \cdot \bm{d}_i),
\end{equation}
where $\bm{n}_z = (0, 0, 1)^T$ is the unit vector along $Z$ axis of the world coordinate system.
Let $\bm{Y}_{neg} = (0, -1, 0)^T$ be the vector of negative $Y$-coordinates of the world coordinate system.
The angle between $\bm{Y}_{neg}$  and $\bm{d}^{proj}_{i}$ is calculated by
\begin{equation}
    \phi = \arctan2(\sin{\phi}, \cos{\phi}),
\end{equation}
where
\begin{eqnarray}
    \cos{\phi} &=& \frac{\bm{d}^{proj}_{i} \cdot \bm{Y}_{neg}}{\| \bm{Y}_{neg} \| \| \bm{d}^{proj}_i \|}, \\
    \sin{\phi} &=& \text{sign}((\bm{Y}_{neg} \times \bm{d}^{proj}_i) \cdot \bm{n}_z) \frac{\| \bm{Y}_{neg} \times \bm{d}^{proj}_{i} \|}{\| \bm{Y}_{neg} \| \| \bm{d}^{proj}_i \|}.
\end{eqnarray}
Using the obtained rotation angle $\phi$ around the $Z$-axis, $R_z(\phi)$ is applied to the world coordinates to rotate the camera coordinate system around the $Z$-axis.
Next, a rotation matrix $R_x(\theta)$ is applied to the camera coordinate system to rotate the camera coordinate system by $\bm{d}^{proj}_{i}$ around the $X$ axis to adjust the $Z$ axis of the rotated camera coordinate system to face the opposite direction of $\bm{d}^{proj}_{i}$ in the world coordinate system.
Finally, the rotation matrix $\bm{R}_i$ of camera $C_i$ from the world coordinate system to the camera coordinate system is obtained as follows.
\begin{equation}
    \bm{R}_i = R_x(\theta)R_z(\phi),
\end{equation}
where
\begin{eqnarray}
    R_x(\theta) &=& 
    \begin{bmatrix}
        1 & 0 & 0 \\
        0 & \sin\theta & -\sin\theta \\
        0 & \cos\theta & \cos\theta
    \end{bmatrix}, \\
    R_z(\phi) &=&
    \begin{bmatrix}
        \cos\phi & \sin\phi & 0 \\
        -\sin\phi & \cos\phi & 0 \\
        0 & 0 & 1 
    \end{bmatrix}.
\end{eqnarray}

\section{Details of The Experiment}

The details of the experiment in this paper are summarized in this section.

\subsection{Computational Resources}

We describe the details of the computational resources used in the experiment in this paper.
The computational resources used in the experiment are summarized in Table \ref{tab:comp-resource}, and the libraries utilized are listed in Table \ref{tab:library-environment}.
For camera parameter estimation, Computer A was used to execute OpenMVG (ver. 2.1.0) and OpenSfM (ver. 0.5.2).
For novel view synthesis, Computer B was used to execute all baseline methods.
For 3D reconstruction, Computer B was also used to execute COLMAP \cite{SchonbergerZPF16}.
The execution environment was set up according to the official COLMAP documentation\footnote{\url{https://colmap.github.io/install.html}}, which provides guidance on installing the necessary libraries.
The execution of NeuS* was conducted on Computer A, while OmniSDF was on Computer C.
In the experiment, Anaconda was utilized to create virtual environments following the setup instructions provided by each method.
The main libraries used include Python, PyTorch, CUDA, and OpenCV, and their versions for each method are detailed in Table \ref{tab:library-environment}.

\begin{table}[t]
    \caption{Computational resources used in the experiments.}
    \label{tab:comp-resource}
    \centering
    \resizebox{\textwidth}{!}{
        \begin{tabular}{rlll}
            \toprule
             Computer & A & B & C \\
            \cmidrule(rl){1-4}
             OS & Ubuntu 22.04.5 & Ubuntu 20.04.04 & Ubuntu 20.04.04 \\
             Memory & 512 GB & 64 GB & 512 GB \\
             CPU & AMD EPYC 7313P 16-Core Processor & Intel(R) Core(TM) i7-6700K CPU @ 4.00GHz & AMD EPYC 7502 32-Core Processor \\
             GPU & NVIDIA GeForce RTX 4090 (24GB) & --- & NVIDIA A100 (80GB) \\
            \bottomrule
        \end{tabular}
    }
\end{table}
\begin{table}[t]
    \caption{The versions of libraries used in the execution of each method.}
    \label{tab:library-environment}
    \centering
    \begin{tabular}{rllll}
      \toprule
         & \multicolumn{3}{c}{Novel View Synthesis} & 3D Reconstruction \\
      \cmidrule(rl){2-4}
      \cmidrule(l){5-5}
        Library & ODGS & op43dgs & EgoNeRF, OmniGS & NeuS*, OmniSDF \\
      \cmidrule(rl){1-1}
      \cmidrule(rl){2-5}
        Python & 3.10.15 & 3.7.12 & 3.9.18 & 3.9.18 \\
        PyTorch & 2.1.2 & 1.12.1 & 2.2.0 & 2.2.0 \\
        CUDA & 11.8 & 11.6 & 12.1 & 12.1 \\
        OpenCV & --- & --- & 4.9.0.80 & 4.9.0.80 \\
        PyTorch3D & --- & --- & 0.7.8 & --- \\
      \bottomrule
    \end{tabular}
\end{table}

\subsection{Camera Parameter Estimation}

We describe the details of the experiment on the baseline methods following on the evaluation protocol of camera parameter estimation for OB3D as shown in Sect. \ref{sec:exp_cpe}.

\subsubsection{Experimental Settings}
\label{sec:cpe_set}

In this experiment, 100 multi-view omnidirectional images are used for each scene, and camera parameters are estimated using SfM.
We summarize each baseline method and the details of the experimental settings.

\noindent
{\bf OpenSfM}\footref{fot:opensfm} ---
OpenSfM is an SfM software supporting omnidirectional images.
OpenSfM employs an incremental strategy where the estimation and optimization of camera parameters are repeated by adding new images sequentially, based on an initial two-view model.
In the experiment, we set ``projection\_type'' to ``spherical'', ``width'' to ``1600'', ``height'' to 800 for all cameras.
The other hyperparameters are the default settings of OpenSfM.

{\bf OpenMVG}\footref{fot:openmvg} ---
OpenMVG is an SfM software supporting omnidirectional images.
OpenMVG estimates the camera parameters based on the global strategy that optimizes the camera parameters for all viewpoints in parallel.
The hyperparameters in our experiments are described below.
First, the ``describerPreset'' is set to ``ULTRA'' in the ``openMVG\_main\_ComputeFeatures'' section, which extracts image feature points.
This setting significantly increases the number of feature points extracted per image, since low-contrast feature points are used and the input image is enlarged by a factor of 2 before feature point extraction is performed.
The ``ratio'' is set to ``0.3'' in ``openMVG\_main\_ComputeMatches'', which is used to match feature points.
The ``ratio'' is a threshold that determines whether the most similar feature point is sufficiently distinct from the next most similar feature point found for a given feature point.
The closer the value is to 1, the less reliable the correspondence is.
The official default setting of 0.8 did not work for camera parameter estimation, so it is adjusted to a lower value and the corresponding points with higher confidence are used.
The other hyperparameters are set to the default settings of the official OpenMVG.

\subsubsection{Evaluation Metrics}

The camera parameters estimated by each method cannot be directly compared with the OB3D camera pose since the scale and position are different from those of the OB3D camera pose.
Therefore, we evaluate the estimation error using the relative camera pose between two arbitrary views, rather than the error for each individual view.
Relative Rotation Error (RRA) and Relative Translation Error (RTA), which represent the angular error between the rotation matrix and the translational vector, are used to evaluate the relative camera parameters.
AUC@5, which indicates the percentage of all image pairs in which both RRA and RTA are less than 5 degrees, is also used as an evaluation metric.
The trajectory gap can be evaluated by matching the estimated camera trajectory with the scale and position of the OB3D camera trajectory.
To evaluate the accuracy of the camera trajectory, the scale and the position of the two camera trajectories are aligned and the Absolute Trajectory Error (ATE), which indicates the Root Mean Squared Error (RMSE) between them, is used.

\subsubsection{Experimental Results}

Table \ref{tab:quantitabe-result-full} shows the experimental results of OpenSfM and OpenMVG.
In indoor scenes, both methods can estimate values close to the camera parameters provided by OB3D, regardless of the camera trajectory.
In outdoor scenes, both methods can estimate the camera parameters with high accuracy for Egocentric trajectories, but for Non-Egocentric trajectories, the estimation accuracy decreased significantly in certain scenes.
In particular, {\tt sponza}, where the accuracy was significantly reduced, has a symmetrical structure compared to other scenes.
This may be due to the fact that the images of {\tt sponza} taken in a Non-Egocentric trajectory have a smaller apparent change in the view, since there is less vertical shift of the view in a Non-Egocentric trajectory.
In OpenSfM, the estimation accuracy of the Non-Egocentric trajectory of {\tt sponza} differs little from other scenes, making it difficult for the global strategy of OpenMVG to deal with images with small apparent changes.

\begin{table}[t]
    \centering
    \caption{Experimental results of camera parameter estimation.}
    \label{tab:quantitabe-result-full}
    \resizebox{\textwidth}{!}{
        \begin{tabular}{ccccccccccc}
        \hline
             & & & \multicolumn{4}{c}{Egocentric} & \multicolumn{4}{c}{Non-Egocentric} \\
            \cmidrule(rl){4-7}
            \cmidrule(rl){8-11}
            Type & Scene & Method & RRA$\downarrow$ & RTA$\downarrow$ & AUC@5$\uparrow$ & ATE$\downarrow$ & RRA$\downarrow$ & RTA$\downarrow$ & AUC@5$\uparrow$ & ATE$\downarrow$\\
            \cmidrule(rl){1-3}
            \cmidrule(rl){4-7}
            \cmidrule(rl){8-11}
            \multirow{12}{*}{\rotatebox{90}{Indoor}} & \multirow{2}{*}{\tt archiviz-flat}  & OpenSfM & 0.405 & 0.044 & 1.000 & 0.000 & 0.406 & 0.059 & 1.000 & 0.000 \\
                                     & & OpenMVG & 0.406 & 0.091 & 1.000 & 0.000 & 0.410 & 0.114 & 0.998 & 0.000\\
            \cmidrule(rl){2-3}
            \cmidrule(rl){4-7}
            \cmidrule(rl){8-11}
            & \multirow{2}{*}{\tt barbershop} & OpenSfM & 0.405 & 0.020 & 1.000 & 0.000 & 0.405 & 0.010 & 1.000 & 0.000 \\
                                     & & OpenMVG & 0.405 & 0.037 & 1.000 & 0.000 & 0.412 & 0.039 & 1.000 & 0.000 \\
            \cmidrule(rl){2-3}
            \cmidrule(rl){4-7}
            \cmidrule(rl){8-11}
            & \multirow{2}{*}{\tt classroom}       & OpenSfM & 0.405 & 0.040 & 1.000 & 0.000 & 0.405 & 0.015 & 1.000 & 0.000 \\
                                     & & OpenMVG & 0.405 & 0.044 & 1.000 & 0.000 & 0.414 & 0.030 & 1.000 & 0.000 \\
            \cmidrule(rl){2-3}
            \cmidrule(rl){4-7}
            \cmidrule(rl){8-11}
            & \multirow{2}{*}{\tt restroom}        & OpenSfM & 0.405 & 0.030 & 1.000 & 0.000 & 0.405 & 0.011 & 1.000 & 0.000 \\
                                     & & OpenMVG & 0.405 & 0.047 & 1.000 & 0.000 & 0.419 & 0.051 & 1.000 & 0.001 \\
            \cmidrule(rl){2-3}
            \cmidrule(rl){4-7}
            \cmidrule(rl){8-11}
            & \multirow{2}{*}{\tt sun-temple}      & OpenSfM & 0.405 & 0.053 & 1.000 & 0.000 & 0.405 & 0.016 & 1.000 & 0.000 \\
                                     & & OpenMVG & 0.405 & 0.107 & 0.999 & 0.000 & 0.408 & 0.056 & 1.000 & 0.000 \\ 
            \cmidrule(rl){2-3}
            \cmidrule(rl){4-7}
            \cmidrule(rl){8-11}
            & \multirow{2}{*}{Average (Indoor)}      & OpenSfM & 0.405 & 0.037 & 1.000 & 0.000 & 0.405 & 0.022 & 1.000 & 0.000 \\
                                     & & OpenMVG & 0.405 & 0.065 & 1.000 & 0.000 & 0.413 & 0.058 & 1.000 & 0.000\\
            \cmidrule(rl){1-11}
            \multirow{16}{*}{\rotatebox{90}{Outdoor}} & \multirow{2}{*}{\tt bistro} & OpenSfM & 0.405 & 0.074 & 1.000 & 0.000 & 0.405 & 0.051 & 1.000 & 0.001 \\
                                     & & OpenMVG & 0.405 & 0.083 & 1.000 & 0.000 & 0.406 & 0.024 & 1.000 & 0.000 \\
            \cmidrule(rl){2-3}
            \cmidrule(rl){4-7}
            \cmidrule(rl){8-11}
            & \multirow{2}{*}{\tt emerald-square}  & OpenSfM & 0.405 & 0.041 & 1.000 & 0.000 & 0.405 & 0.024 & 1.000 & 0.000 \\
                                     & & OpenMVG & 0.405 & 0.106 & 1.000 & 0.000 & 0.405 & 0.064 & 1.000 & 0.001 \\
            \cmidrule(rl){2-3}
            \cmidrule(rl){4-7}
            \cmidrule(rl){8-11}
            & \multirow{2}{*}{\tt fisher-hut}      & OpenSfM & 0.405 & 0.027 & 1.000 & 0.000 & 0.405 & 0.022 & 1.000 & 0.001 \\
                                     & & OpenMVG & 0.405 & 0.120 & 0.999 & 0.000 & 0.406 & 0.122 & 0.998 & 0.002 \\
            \cmidrule(rl){2-3}
            \cmidrule(rl){4-7}
            \cmidrule(rl){8-11}
            & \multirow{2}{*}{\tt lone-monk}       & OpenSfM & 0.405 & 0.018 & 1.000 & 0.000 & 0.405 & 0.012 & 1.000 & 0.000 \\
                                    & & OpenMVG & 0.405 & 0.062 & 1.000 & 0.000 & 0.408 & 0.028 & 1.000 & 0.000 \\
            \cmidrule(rl){2-3}
            \cmidrule(rl){4-7}
            \cmidrule(rl){8-11}
            & \multirow{2}{*}{\tt pavillion}       & OpenSfM & 0.405 & 0.033 & 1.000 & 0.000 & 0.410 & 0.074 & 1.000 & 0.001 \\
                                     & & OpenMVG & 0.405 & 0.080 & 1.000 & 0.000 & 0.411 & 0.099 & 1.000 & 0.001 \\
            \cmidrule(rl){2-3}
            \cmidrule(rl){4-7}
            \cmidrule(rl){8-11}
            & \multirow{2}{*}{\tt san-miguel}      & OpenSfM & 0.405 & 0.023 & 1.000 & 0.000 & 0.405 & 0.011 & 1.000 & 0.000 \\
                                     & & OpenMVG & 0.405 & 0.070 & 1.000 & 0.000 & 0.427 & 2.544 & 0.960 & 0.089 \\
            \cmidrule(rl){2-3}
            \cmidrule(rl){4-7}
            \cmidrule(rl){8-11}
            & \multirow{2}{*}{\tt sponza}          & OpenSfM & 0.405 & 0.012 & 1.000 & 0.000 & 0.405 & 0.009 & 1.000 & 0.000 \\
                                      & & OpenMVG & 0.405 & 0.014 & 1.000 & 0.000 & 84.239 & 42.221 & 0.000 & 0.542 \\
            \cmidrule(rl){2-3}
            \cmidrule(rl){4-7}
            \cmidrule(rl){8-11}
            & \multirow{2}{*}{Average (Outdoor)}     & OpenSfM & 0.405 & 0.033 & 1.000 & 0.000 & 0.406 & 0.033 & 1.000 & 0.000 \\
                                     & & OpenMVG & 0.405 & 0.077 & 1.000 & 0.000 & 12.386 & 6.443 & 0.851 & 0.091 \\
            \cmidrule(rl){1-11}
            & \multirow{2}{*}{Average (All Scenes)}  & OpenSfM & 0.405 & 0.035 & 1.000 & 0.000 & 0.406 & 0.026 & 1.000 & 0.000 \\
                                     & & OpenMVG & 0.405 & 0.072 & 1.000 & 0.000 & 7.397 & 3.783 & 0.913 & 0.053 \\  
        \hline
        \end{tabular}
    }
\end{table}

\subsubsection{Evaluation of Camera Parameter Estimation by Changing the Number of Views}

\begin{figure}[t]
     \centering
     \includegraphics[width=\linewidth]{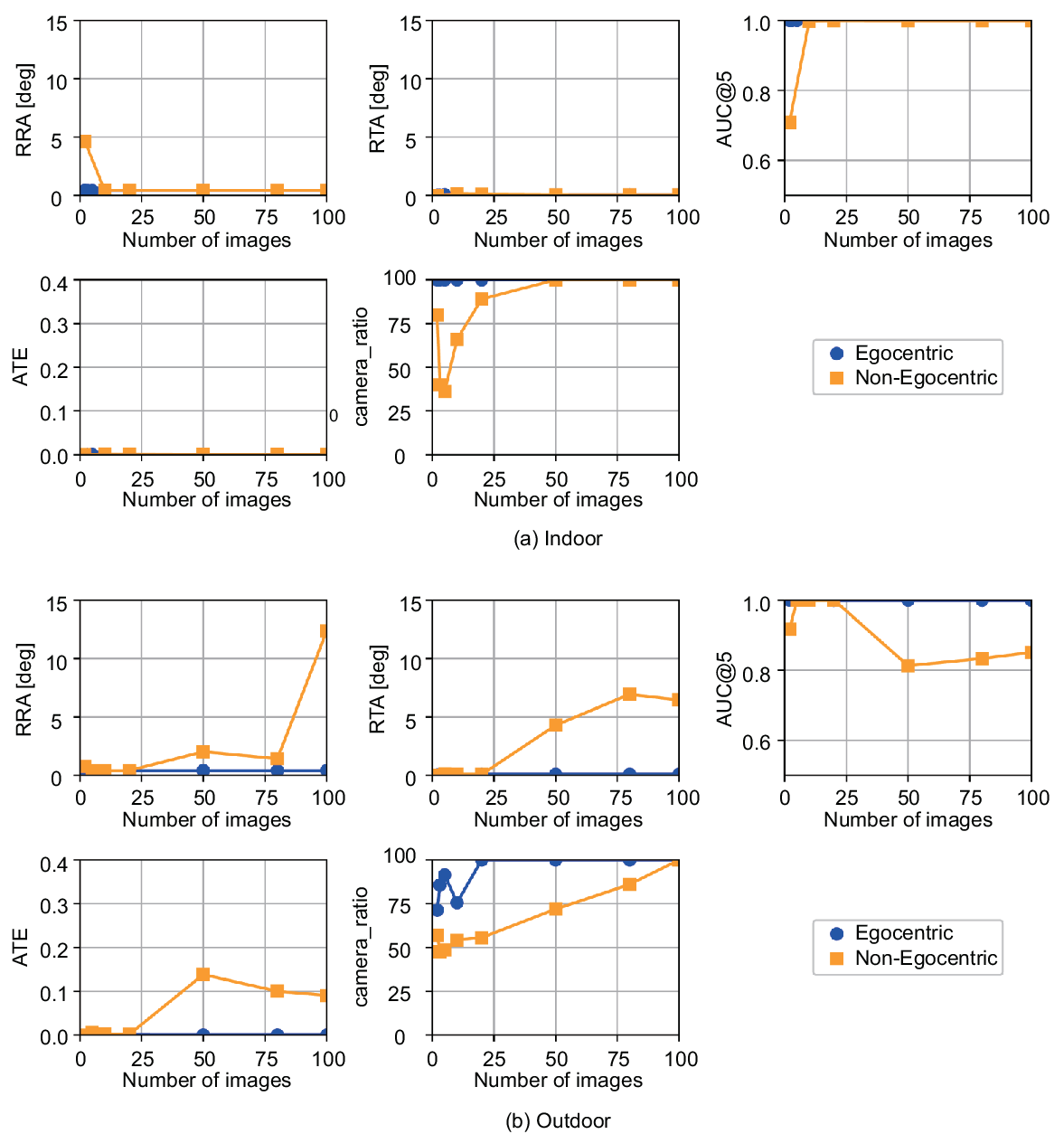}
     \caption{Experimental results of camera parameter estimation using OpenMVG when changing the number of views to be input.}
     \label{fig:supp_imgs_all}
 \end{figure}

In this experiment, the number of viewpoints used for camera parameter estimation is changed to 2, 3, 5, 10, 20, 50, 80, and 100, and the change in estimation accuracy is investigated.
RRA, RTA, AUC@5, and ATE are used as evaluation metrics, and the change in camera\_ratio, the ratio of views for which camera parameters are estimated out of the input views, is also analyzed.
Note that if the number of input views and the number of estimated camera parameters are not equal, the evaluation metrics for the camera parameters are not calculated.
In this experiment, OpenMVG is used as the baseline.
Fig. \ref{fig:supp_imgs_all} shows the average of each metric in the indoor and outdoor scenes.
In both scenes, the camera parameters can be estimated more accurately with fewer views using the Egocentric trajectory.
In the indoor scene, the camera parameters can be estimated for all views with about 50 views, regardless of the camera trajectory.
On the other hand, in the case of the outdoor scene with the Egocentric trajectory, the camera parameters can be estimated with only about 20 views, while about 100 views are required in the case of a Non-Egocentric trajectory.
Furthermore, in the case of the Non-Egocentric trajectory for the outdoor scene, increasing the number of views to 100 significantly reduces the accuracy of RRA.
This is due to the fact that the accuracy of camera parameter estimation for {\tt sponza} drops drastically when the number of views is 100, as shown in Table \ref{tab:quantitabe-result-full}.
However, it is possible to estimate the camera parameters for {\tt sponza} with relatively high accuracy when the number of views is around 80.

\subsection{Novel View Synthesis}

We describe the details of the experiments on the baseline methods following on the evaluation protocol of novel view synthesis for OB3D as shown in Sect. \ref{sec:exp_nvs}.

\subsubsection{Experimental Settings}

In this experiment, the radiance field is trained using images in the train dataset, a novel view image is rendered based on the radiance field and the view of the test data, and the error between the test data image and the rendered image is evaluated.
Then, PSNR, SSIM, LPIPS (A), and LPIPS (V) of the rendered images are used to evaluate the accuracy of novel view synthesis.
We summarize each baseline method and the details of the experimental settings.

\noindent
{\bf EgoNeRF} \cite{ChoiK023} ---
EgoNeRF optimizes the radiance field in a wide-area space using images taken in an Egocentric trajectory as input, and performs novel view synthesis.
Instead of the Cartesian grid used in the conventional methods, two balanced spherical grids that take into account the camera model of omnidirectional images are introduced as feature grids to represent the radiance field to improve the rendering accuracy.
In this experiment, the official implementation of EgoNeRF\footnote{\url{https://github.com/changwoonchoi/EgoNeRF}} is used, except that the data loader is modified to conduct the experiment in OB3D.
The hyperparameters are the same as in the official implementation of EgoNeRF.
The number of iterations for optimization is set to 50,000.

\noindent
{\bf ODGS} \cite{LeeCHL24} ---
ODGS is a method that modifies part of the rasterization of 3DGS so that the camera model of the omnidirectional image is the unit sphere and 3D Gaussians are projected onto the surface of the sphere.
The distortion inherent in omnidirectional images is taken into account when densifying the 3D Gaussians to improve the accuracy of the novel view synthesis.
In this experiment, the official ODGS implementation\footnote{\url{https://github.com/esw0116/ODGS}} is used, except that the data loader is modified to conduct the experiments in OB3D.
The hyperparameters are the same as in the official ODGS implementation.
The number of iterations of optimization is set to 30,000.

\noindent
{\bf op43dgs} \cite{HuangBGLG24} ---
op43dgs is a method that enables 3DGS to be performed on any camera model by deriving the optimal projection method for 3DGS through the analysis of the minimum value of the projection function based on functional optimization theory.
In this experiment, the official implementation of op43dgs\footnote{\url{https://github.com/LetianHuang/op43dgs}} is used, except that the data loader is modified to conduct the experiment in OB3D.
The hyperparameters are the same as in the official op43dgs implementation.
The number of iterations of optimization is set to 30,000.

\noindent
{\bf OmniGS} \cite{LiHYC24} ---
Similar to ODGS \cite{LeeCHL24}, OmniGS is a method that modifies part of the rasterization of 3DGS so that the camera model of the omnidirectional image is the unit sphere and 3D Gaussians are projected onto the surface of the sphere.
Noise is suppressed by removing 3D Gaussians that are too large during optimization.
Although the official implementation of OmniGS\footnote{\url{https://github.com/liquorleaf/OmniGS}} is available, it is written in C++, so we used our reproduction in Python for this experiment.
The hyperparameters are the same as in the official OmniGS implementation.
The number of iterations of optimization is set to 30,000.

\subsubsection{Evaluation Metrics}

In this experiment, PSNR [d], SSIM [e], and LPIPS [f] are used as evaluation metrics for novel view synthesis.
PSNR is a metric that evaluates image quality degradation based on the ratio of the maximum power of the signal to the noise that affects image quality.
SSIM is a metric that evaluates the similarity of the structural information of objects in an image, rather than the visual similarity recognizable by humans and animals, with a similarity close to 1 corresponding to higher image quality.
LPIPS is a metric that evaluates image quality based on the distance between the resulting feature maps extracted by the deep learning model.
In this experiment, LPIPS (A) using AlexNet and LPIPS (V) using VGG as deep learning models are used to evaluate accuracy.

\subsubsection{Experimental Results}

Table \ref{tab:nvs-quantitative-all} shows the results of the experiment with the LPIPS(V) evaluation results added to Table \ref{tab:nvs-quantitative-result}.
EgoNeRF shows high rendering accuracy in both environments and is robust to scene changes.
The 3DGS-based methods vary their rendering accuracy for the environment changes from indoor to outdoor, and for the camera trajectory changes from Egocentric to Non-Egocentric.
In particular, op43dgs exhibits the same level of accuracy as the other methods for outdoor Non-Egocentric images, however, the variation of accuracy is large for changes in environment and trajectory.

\begin{figure}[t]
    \centering
    \includegraphics[width=\linewidth]{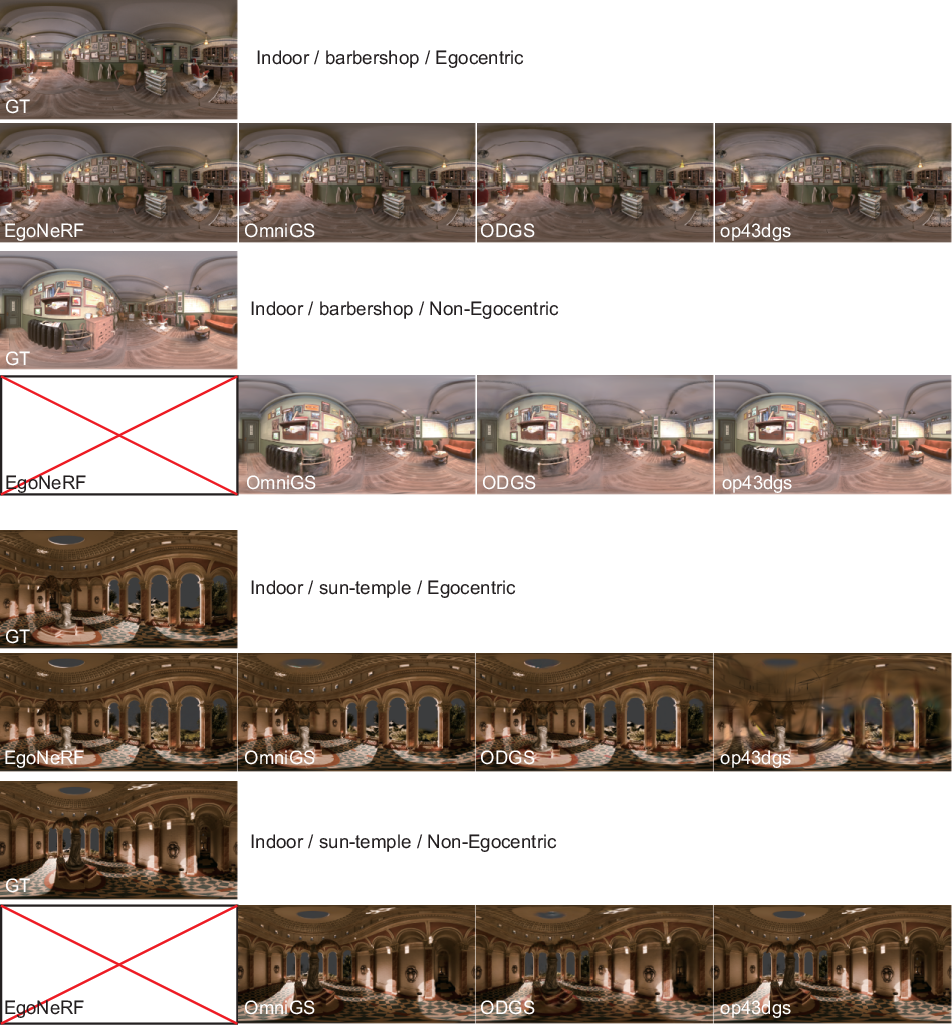}
    \caption{Examples of rendered novel views for some indoor scenes.}
    \label{fig:nvs_qualitative_indoor}
\end{figure}

\begin{figure}[t]
    \centering
    \includegraphics[width=\linewidth]{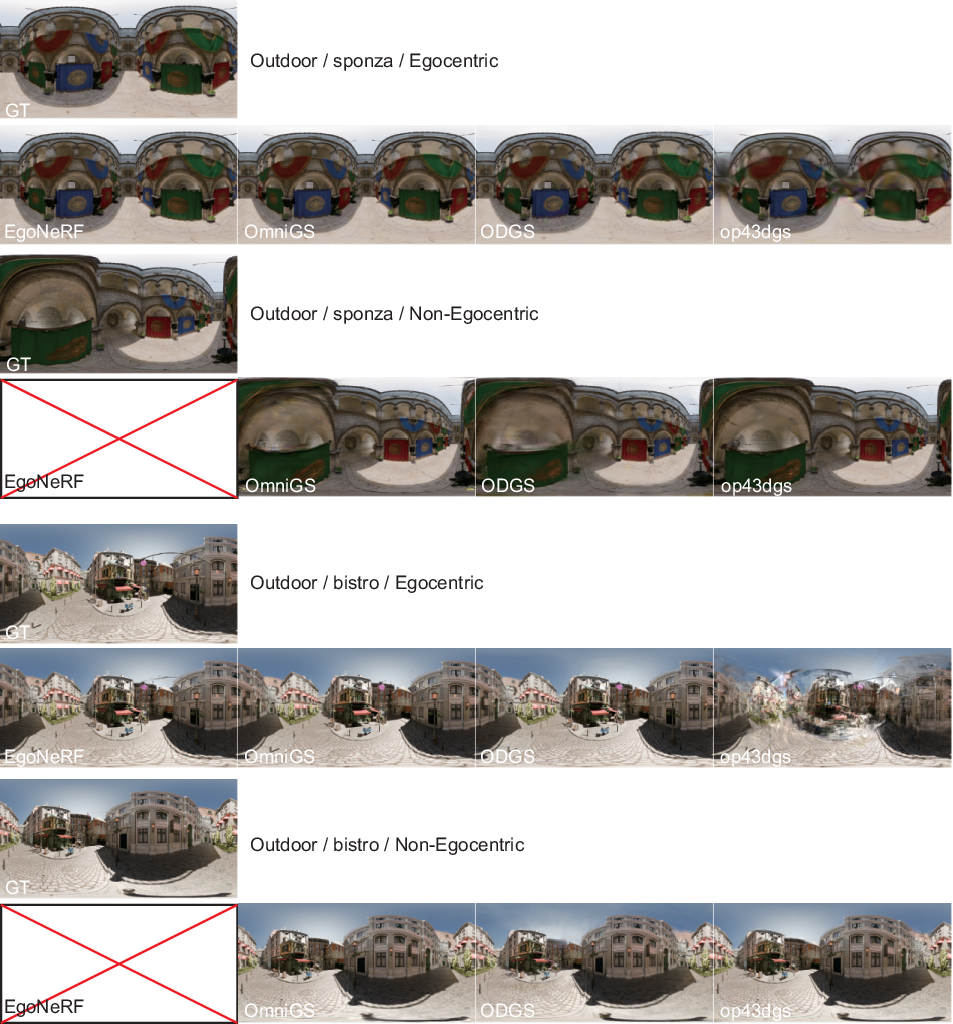}
    \caption{Examples of rendered novel views for some outdoor scenes.}
    \label{fig:nvs_qualitative_outdoor}
\end{figure}

Figs. \ref{fig:nvs_qualitative_indoor} and \ref{fig:nvs_qualitative_outdoor} show the qualitative evaluation results of novel view synthesis using each method for indoor and outdoor scenes in OB3D.
Note that EgoNeRF assumes the input of Egocentric images and is not included in the experimental results for Non-Egocentric images.
The accuracy of rendering in different environments varied depending on the method used.
In particular, a comparison of the rendering accuracy of the different methods for Egocentric and Non-Egocentric images shows that ODGS and op43dgs render very differently visually, while OmniGS renders better than op43dgs.
On the other hand, OmniGS keeps a stable rendering accuracy for both Egocentric and Non-Egocentric cases.
ODGS may also be specialized to handle images under special capture conditions, such as Egocentric images.
In contrast, quantitative and qualitative evaluation results suggest that op43dgs is better suited for novel view synthesis using standard capture strategies, such as Non-Egocentric images.

\begin{table}[t]
    \centering
    \caption{Experimental results of novel view synthesis in all scenes, where the values indicate the average of each metric.
    Note that EgoNeRF is not applicable to Non-Egocentric images since it assumes that Egocentric images are input.}
    \label{tab:nvs-quantitative-all}
    \resizebox{\textwidth}{!}{
    \begin{tabular}{cccccccccc}
         \toprule
          & & \multicolumn{4}{c}{Egocentric} & \multicolumn{4}{c}{Non-Egocentric} \\
         \cmidrule(rl){3-6}
         \cmidrule(rl){7-10}
          Type & Method & PSNR [dB] $\uparrow$ & SSIM $\uparrow$ & LPIPS (A) $\downarrow$ & LPIPS (V) $\downarrow$ & PSNR [dB] $\uparrow$ & SSIM $\uparrow$ & LPIPS (A) $\downarrow$ & LPIPS (V) $\downarrow$ \\
         \cmidrule(rl){1-2}
         \cmidrule(l){3-10}
            \multirow{4}{*}{\rotatebox{90}{Indoor}} & EgoNeRF \cite{ChoiK023} 
             & 32.26 & {\bf 0.906} & {\bf 0.128} & {\bf 0.220} & --- & --- & --- & --- \\
            & ODGS \cite{LeeCHL24} 
             & 28.14 & 0.840 & 0.241 & 0.373 & 26.80 & 0.819 & 0.256 & 0.289 \\
            & op43dgs \cite{HuangBGLG24} 
             & 23.86 & 0.780 & 0.402 & 0.441 & {\bf 31.18} & {\bf 0.905} & {\bf 0.154} & 0.441 \\
            & OmniGS \cite{LiHYC24} 
             & {\bf 33.25} & 0.897 & 0.169 & 0.276 & 27.93 & 0.839 & 0.247 & {\bf 0.183} \\
         \cmidrule(rl){1-2}
         \cmidrule(l){3-10}
           \multirow{4}{*}{\rotatebox{90}{Outdoor}} & EgoNeRF \cite{ChoiK023} 
            & {\bf 32.18} & {\bf 0.916} & {\bf 0.086} & {\bf 0.165} & --- & --- & --- & --- \\
             & ODGS \cite{LeeCHL24} 
            & 27.47 & 0.849 & 0.176 & 0.399 & 25.04 & 0.771 & {\bf 0.243} & 0.365 \\
             & op43dgs \cite{HuangBGLG24} 
            & 19.22 & 0.685 & 0.407 & 0.262 & 25.11 & {\bf 0.828} & 0.244 & {\bf 0.314} \\
             & OmniGS \cite{LiHYC24} 
            & 29.89 & 0.907 & 0.112 & 0.348 & {\bf 25.86} & 0.785 & 0.258 & 0.324 \\
         \bottomrule
    \end{tabular}
    }
\end{table}


\subsection{3D Reconstruction}

We describe the details of experiments on the baseline methods following on the evaluation protocol of 3D reconstruction for OB3D as shown in Sect. \ref{sec:exp_3dr}.

\subsubsection{Experimental Settings}

In this experiment, a 3D mesh model is reconstructed from images in the train dataset, and a depth map is generated by rendering the mesh model from the same viewpoint as the images in the train dataset.
The accuracy of the 3D reconstruction is evaluated by calculating the error between the depth map rendered from the mesh model and the depth map provided by OB3D.
We summarize each baseline method and the details of the experimental settings.

{\bf COLMAP} \cite{SchonbergerZPF16} --- 
COLMAP is a 3D reconstruction method that performs matching between multi-view images, estimates the depth map of each viewpoint, and integrates them.
Since COLMAP assumes perspective projection images as input, it cannot perform 3D reconstruction directly from omnidirectional images.
Therefore, in this experiment, omnidirectional images are preprocessed to create cube maps, which are perspective projection images from six different views, i.e., front, back, right, left, top, and bottom, and these are used as input images for COLMAP.

{\bf NeuS} \cite{WangLLTKW21} --- 
NeuS is a 3D reconstruction method that combines an implicit function representation of the scene and optimization of the radiance field.
Through volume rendering using the SDF and radiance field, NeuS generates an image from the same view as the image in the train dataset and optimizes it to minimize the error from the ground-truth image.
After completing the optimization, the 3D points with an SDF value of 0 are extracted and the mesh model is reconstructed by applying the marching cubes algorithm \cite{LorensenC87}.
NeuS does not support rendering of omnidirectional images since it assumes perspective projection images as input.
In this experiment, we modified NeuS\footnote{\url{https://github.com/ShntrIto/SDF360/tree/main}} to be able to render omnidirectional images by changing the rays for perspective projection images to those for omnidirectional images, which serves as a baseline.
We optimize to minimize the error between the rendered omnidirectional images and the ground-truth omnidirectional images, and use the resulting SDF to perform 3D reconstruction directly from the omnidirectional images.
The hyperparameters are the same as in the official implementation of NeuS\footnote{\url{https://github.com/Totoro97/NeuS}}.
The number of iterations of optimization is set to 200,000.

{\bf OmniSDF} \cite{KimMJTK24} --- 
OmniSDF is a method for reconstructing the surface of objects by estimating the spatial SDF based on NeuS, using multi-view omnidirectional images, corresponding camera parameters, and depth maps as input.
When performing 3D reconstruction of a wide-area space, many regions do not contain any objects.
OmniSDF uses a binoctree to determine the space in which no objects exist, and performs ray sampling limited to the grid in which objects exist, thereby achieving highly efficient SDF estimation for wide-area spaces.
Since OmniSDF uses a depth map for initializing the binoctree, the depth map included in OB3D is used for initializing the binoctree in this experiment.
In this experiment, the official implementation of OmniSDF\footnote{\url{https://github.com/KAIST-VCLAB/OmniSDF}} is used, except that the data loader is modified to conduct the experiment in OB3D.
The hyperparameters are the same as in the official OmniSDF implementation.
The number of iterations of optimization is set to 200,000.

\subsubsection{Evaluation Metrics}

The accuracy of the mesh models reconstructed by each method is evaluated using the Root Mean Squared Error (RMSE), Mean Absolute Error (MAE), Absolute Relative Error (AbsRel) and $\delta_{1.25}$ for the depth map rendered.
AbsRel is a metric that evaluates the absolute relative error between the rendered depth and the ground-truth depth.
AbsRel is defined as
\begin{equation}
    \text{AbsRel} = \frac{1}{N} \sum_{i=1}^{N} \frac{\left| z_i - z_i^\ast \right|}{z_i^\ast},
\end{equation}
where $z_i$ is the depth rendered from the mesh model, $z_i^\ast$ is the ground-truth depth, $i$ is the pixel index, and $N$ is the total number of pixels.
Lower values of AbsRel indicate higher reconstruction accuracy.
$\delta_{1.25}$ is the ratio of the rendered depth to the ground-truth depth of 1.25 or less among all the pixels to be evaluated, with higher values indicating higher reconstruction accuracy.

\subsubsection{Experimental Results}

\begin{figure}
    \centering
    \includegraphics[width=1\linewidth]{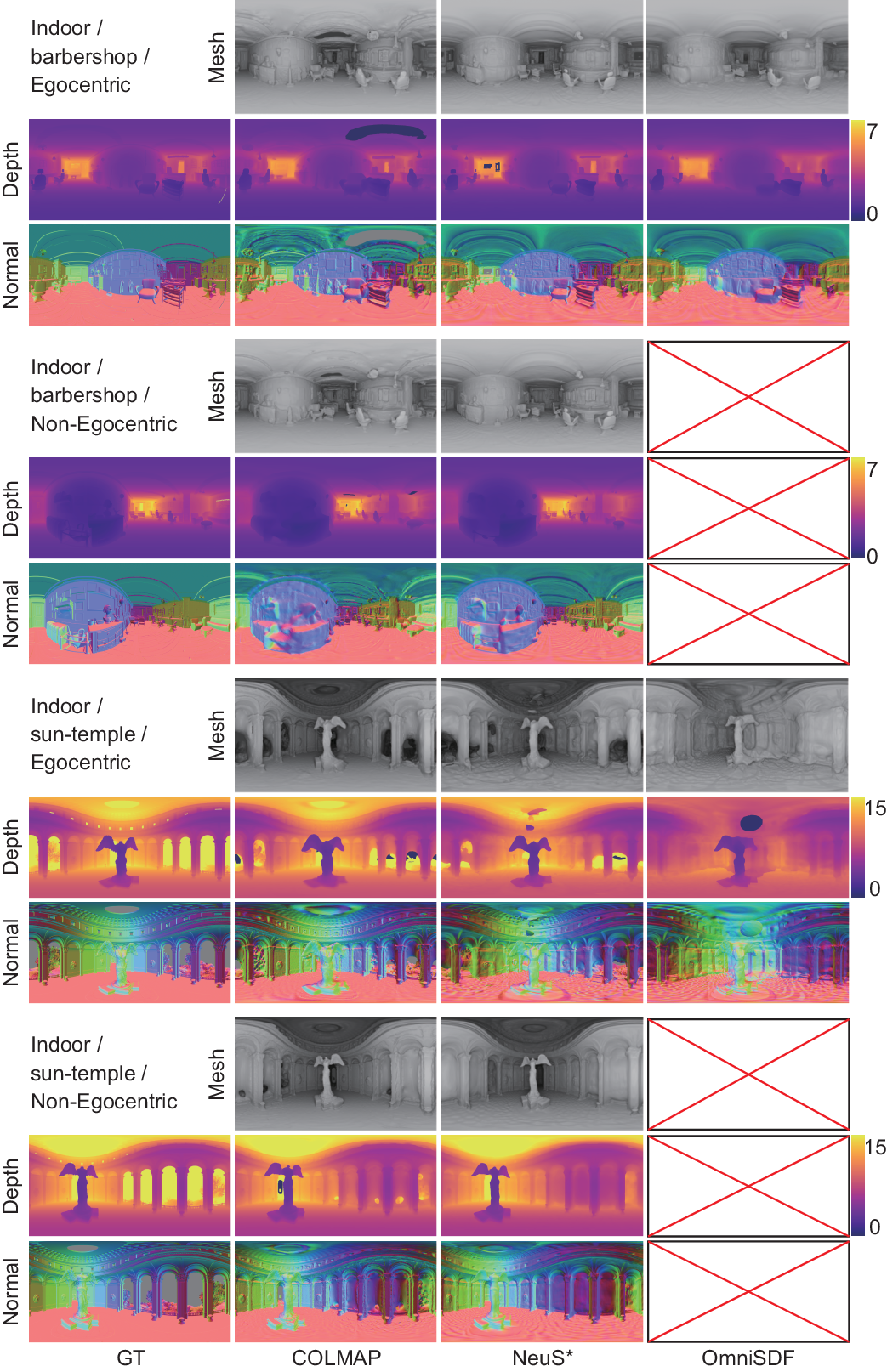}
    \caption{Example of mesh model, depth map, and normal map in 3D reconstruction for the indoor scene.}
    \label{fig:supp-results-3d-reconstruction-indoor}
\end{figure}

\begin{figure}
    \centering
    \includegraphics[width=1\linewidth]{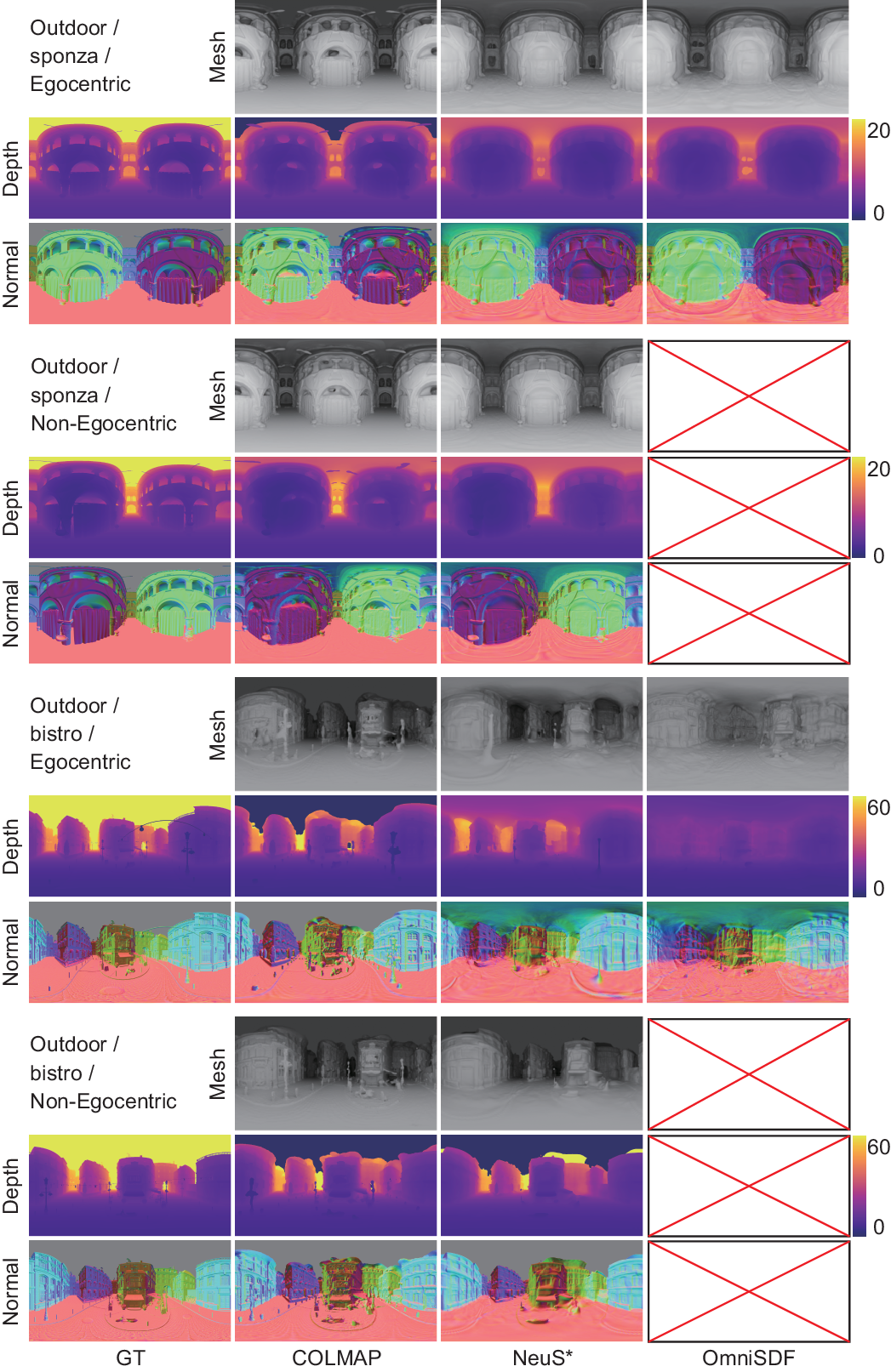}
    \caption{Example of mesh model, depth map, and normal map in 3D reconstruction for the outdoor scene.}
    \label{fig:supp-results-3d-reconstruction-outdoor}
\end{figure}

Tables \ref{tab:supp-quantitative-3d-recontruction-colmap}, \ref{tab:supp-quantitative-3d-recontruction-neus}, and \ref{tab:supp-quantitative-3d-reconstruction-omnisdf} show the summary of experimental results and Figs. \ref{fig:supp-results-3d-reconstruction-indoor}, \ref{fig:supp-results-3d-reconstruction-outdoor} show examples of 3D reconstruction for each baseline method.
Note that experimental results with Non-Egocentric images are not included in OmniSDF, since it assumes Egocentric images as input.
The ``Average (all)'' indicates the average of the ``Average (Indoor)'' and ``Average (Outdoor)'', and ``Average (All scenes)'' indicates the average of all scenes.
From the experimental results, we can confirm that the accuracy of 3D reconstruction varies depending on the scene type and camera trajectory.
Comparing the accuracy of the indoor and outdoor scenes in COLMAP, RMSE is lower in the outdoor scene.
This is due to the larger scale of the outdoor scene, resulting in a larger error.
On the other hand, for AbsRel and $\delta_{1.25}$, which are less sensitive to the scene scale, the accuracy of the indoor scene is lower than that of the outdoor scene.
This is because indoor scenes contain more planar surfaces and poor texture, which makes matching between images more difficult.
Compared to COLMAP, which uses perspective projection images as input, NeuS, which can process omnidirectional images directly, has higher accuracy for both Egocentric and Non-Egocentric trajectories.
Furthermore, a comparison of the accuracy of each scene in NeuS* between Egocentric and Non-Egocentric trajectories shows that the Non-Egocentric trajectory has a higher reconstruction accuracy.
OmniSDF shows high accuracy in some indoor scenes, however, it shows lower accuracy in outdoor scenes, and in some cases, it fails to reconstruct the scene.

\begin{table}[t]
    \centering
    \caption{Experimental results of 3D reconstruction using COLMAP \cite{SchonbergerZPF16}.}
    \label{tab:supp-quantitative-3d-recontruction-colmap}
    \resizebox{\textwidth}{!}{
        \begin{tabular}{cccccccccc}
        \hline
             &  & \multicolumn{4}{c}{Egocentric} & \multicolumn{4}{c}{Non-Egocentric} \\
            \cmidrule(rl){3-6}
            \cmidrule(rl){7-10}
            Type & Scene & RMSE [m] $\downarrow$ & MAE [m] $\downarrow$ & AbsRel$\downarrow$ & $\delta_{1.25}$ [\%] $\uparrow$ & RMSE [m] $\downarrow$ & MAE [m] $\downarrow$ & AbsRel$\downarrow$ & $\delta_{1.25}$ [\%] $\uparrow$ \\
            \cmidrule(rl){1-2}
            \cmidrule(rl){3-6}
            \cmidrule(rl){7-10}
            \multirow{6}{*}{\rotatebox{90}{Indoor}} & {\tt archiviz-flat} & 0.883 & 0.502 & 0.308 & 0.688 & 1.073 & 0.639 & 0.359 & 0.635 \\
            & {\tt barbershop}      & 0.277 & 0.079 & 0.041 & 0.960 & 0.240 & 0.073 & 0.043 & 0.950 \\
            & {\tt classroom}       & 0.520 & 0.107 & 0.033 & 0.953 & 0.725 & 0.257 & 0.129 & 0.855 \\
            & {\tt restroom}        & 2.359 & 0.578 & 0.060 & 0.936 & 1.152 & 0.204 & 0.026 & 0.969 \\
            & {\tt sun-temple}      & 0.848 & 0.192 & 0.018 & 0.987 & 0.776 & 0.158 & 0.017 & 0.986 \\ 
            & Average (Indoor)      & 0.978 & 0.292 & 0.092 & 0.905 & 0.793 & 0.266 & 0.115 & 0.879 \\
            \cmidrule(rl){1-2}
            \cmidrule(rl){3-6}
            \cmidrule(rl){7-10}
            \multirow{8}{*}{\rotatebox{90}{Outdoor}} & {\tt bistro} & 1.842 & 0.368 & 0.061 & 0.981 & 1.802 & 0.367 & 0.037 & 0.972 \\
            & {\tt emerald-square}  & 3.738 & 0.621 & 0.023 & 0.979 & 3.869 & 0.649 & 0.025 & 0.975 \\
            & {\tt fisher-hut}      & 3.068 & 0.537 & 0.077 & 0.931 & 2.779 & 0.570 & 0.119 & 0.885 \\
            & {\tt lone-monk}       & 1.664 & 0.409 & 0.042 & 0.928 & 1.316 & 0.370 & 0.046 & 0.912 \\
            & {\tt pavillion}       & 2.281 & 0.481 & 0.041 & 0.963 & 2.515 & 0.591 & 0.060 & 0.967 \\
            & {\tt san-miguel}      & 2.009 & 0.667 & 0.097 & 0.835 & 2.327 & 0.954 & 0.190 & 0.781 \\
            & {\tt sponza}          & 1.065 & 0.308 & 0.042 & 0.933 & 0.984 & 0.296 & 0.049 & 0.922 \\
            & Average (Outdoor)     & 2.238 & 0.485 & 0.055 & 0.936 & 2.228 & 0.542 & 0.075 & 0.916 \\
            \cmidrule(rl){1-2}
            \cmidrule(rl){3-6}
            \cmidrule(rl){7-10}
            & Average (All scenes)  & 1.713 & 0.404 & 0.070 & 0.923 & 1.630 & 0.427 & 0.092 & 0.901 \\ 
        \hline
        \end{tabular}
    }
\end{table}
\begin{table}[t]
    \centering
    \caption{Experimental results of 3D reconstruction using NeuS*.}
    \label{tab:supp-quantitative-3d-recontruction-neus}
    \resizebox{\textwidth}{!}{
        \begin{tabular}{cccccccccc}
        \hline
            & & \multicolumn{4}{c}{Egocentric} & \multicolumn{4}{c}{Non-Egocentric} \\
            \cmidrule(rl){3-6}
            \cmidrule(rl){7-10}            
            Type & Scene & RMSE [m] $\downarrow$ & MAE [m] $\downarrow$ & AbsRel$\downarrow$ & $\delta_{1.25}$ [\%] $\uparrow$ & RMSE [m] $\downarrow$ & MAE [m] $\downarrow$ & AbsRel$\downarrow$ & $\delta_{1.25}$ [\%] $\uparrow$ \\
            \cmidrule(rl){1-2}
            \cmidrule(rl){3-6}
            \cmidrule(rl){7-10}
            \multirow{6}{*}{\rotatebox{90}{Indoor}} & {\tt archiviz-flat} & 0.206 & 0.069 & 0.035 & 0.994 & 0.191 & 0.050 & 0.025 & 0.986 \\
            & {\tt barbershop}      & 0.319 & 0.057 & 0.024 & 0.991 & 0.121 & 0.039 & 0.024 & 0.986 \\
            & {\tt classroom}       & 0.175 & 0.037 & 0.013 & 0.991 & 0.189 & 0.038 & 0.018 & 0.982 \\
            & {\tt restroom}        & 0.423 & 0.147 & 0.021 & 0.989 & 0.499 & 0.142 & 0.023 & 0.978 \\
            & {\tt sun-temple}      & 0.908 & 0.297 & 0.028 & 0.978 & 0.797 & 0.205 & 0.024 & 0.985 \\ 
            & Average (Indoor)      & 0.406 & 0.121 & 0.024 & 0.989 & 0.359 & 0.095 & 0.023 & 0.983 \\
            \cmidrule(rl){1-2}
            \cmidrule(rl){3-6}
            \cmidrule(rl){7-10}
            \multirow{8}{*}{\rotatebox{90}{Outdoor}} & {\tt bistro} & 2.070 & 0.673 & 0.080 & 0.970 & 2.348 & 0.551 & 0.059 & 0.970 \\
            & {\tt emerald-square}  & 6.235 & 1.552 & 0.058 & 0.951 & 3.772 & 0.766 & 0.046 & 0.976 \\
            & {\tt fisher-hut}      & 2.975 & 0.645 & 0.083 & 0.938 & 3.345 & 0.627 & 0.093 & 0.949 \\
            & {\tt lone-monk}       & 1.251 & 0.404 & 0.049 & 0.912 & 1.322 & 0.383 & 0.056 & 0.925 \\
            & {\tt pavillion}       & 2.522 & 0.576 & 0.038 & 0.952 & 2.518 & 0.661 & 0.081 & 0.958 \\
            & {\tt san-miguel}      & 1.357 & 0.564 & 0.118 & 0.826 & 1.187 & 0.428 & 0.091 & 0.872 \\
            & {\tt sponza}          & 1.311 & 0.456 & 0.062 & 0.897 & 1.148 & 0.385 & 0.060 & 0.892 \\
            & Average (Outdoor)     & 2.532 & 0.696 & 0.070 & 0.921 & 2.234 & 0.543 & 0.070 & 0.934 \\
            \cmidrule(rl){1-2}
            \cmidrule(rl){3-6}
            \cmidrule(rl){7-10}
            & Average (All scenes)  & 1.646 & 0.456 & 0.051 & 0.949 & 1.453 & 0.356 & 0.050 & 0.955 \\ 
        \hline
        \end{tabular}
    }
\end{table}
\begin{table}[t]
    \centering
    \caption{Experimental results of 3D reconstruction using OmniSDF \cite{KimMJTK24}.}
    \label{tab:supp-quantitative-3d-reconstruction-omnisdf}
    \begin{tabular}{cccccc}
         \toprule
             & & \multicolumn{4}{c}{Egocentric}\\
         \cmidrule(rl){3-6}
             Type & Scene & RMSE [m] $\downarrow$ & MAE [m] $\downarrow$ & AbsRel $\downarrow$ & $\delta_{1.25}$ [\%] $\uparrow$ \\
         \cmidrule(rl){1-2}
         \cmidrule(rl){3-6}
             \multirow{6}{*}{\rotatebox{90}{Indoor}}
             & {\tt archiviz-flat}  & 0.238 & 0.062 & 0.025 & 0.995 \\
             & {\tt barbershop}     & 0.136 & 0.053 & 0.026 & 0.988 \\
             & {\tt classroom}      & 0.224 & 0.069 & 0.026 & 0.982 \\
             & {\tt restroom}       & 2.235 & 1.147 & 0.139 & 0.753 \\
             & {\tt sun-temple}     & 2.759 & 1.638 & 0.140 & 0.708 \\
             & Average (Indoor)     & 1.118 & 0.594 & 0.071 & 0.885 \\
         \cmidrule(rl){1-2}
         \cmidrule(rl){3-6}
             \multirow{8}{*}{\rotatebox{90}{Outdoor}}
             & {\tt bistro}         & 6.226 & 2.748 & 0.144 & 0.809 \\
             & {\tt emerald-square} & 6.000 & 2.238 & 0.127 & 0.843 \\
             & {\tt fisher-hut}     & 5.698 & 1.428 & 0.099 & 0.884 \\
             & {\tt lone-monk}      & 1.318 & 0.457 & 0.052 & 0.906 \\
             & {\tt pavillion}      & 2.832 & 0.642 & 0.032 & 0.951 \\
             & {\tt san-miguel}     & --- & --- & --- & --- \\
             & {\tt sponza}         & 1.320 & 0.482 & 0.068 & 0.899 \\
             & Average (Outdoor)    & 3.899 & 1.333 & 0.087 & 0.882 \\
         \cmidrule(rl){1-2}
         \cmidrule(rl){3-6}
             & Average (All scenes) & 2.635 & 0.997 & 0.080 & 0.883 \\
         \bottomrule
    \end{tabular}
\end{table}


\section{Additional References}

{\small
\begin{enumerate}
[leftmargin=5mm]
\renewcommand{\labelenumi}{[\alph{enumi}]}
\item A.~Lumberyard, ``Amazon lumberyard bistro, open research content archive ({ORCA}),'' July 2017.
\newblock \small \texttt{http://developer.nvidia.com/orca/amazon-lumberyard-bistro}.
\item N.~Hull, K.~Anderson, and N.~Benty, ``{NVIDIA} emerald square, open research content archive ({ORCA}),'' July 2017.
\newblock \small \texttt{http://developer.nvidia.com/orca/nvidia-emerald-square}.
\item E.~Games, ``Unreal engine sun temple, open research content archive ({ORCA}),'' Oct. 2017.
\newblock \small \texttt{http://developer.nvidia.com/orca/epic-games-sun-temple}.
\item Q.~Huynh-Thu and M.~Ghanbari, ``Scope of validity of {PSNR} in image/video quality assessment,'' {\em Electronics Letters}, vol.~44, pp.~800--801, Feb. 2008.
\item Z.~Wang, A.~Bobik, and E.~Sheikh, H.R.~Simoncelli, ``Image quality assessment: {F}rom error visibility to structural similarity,'' {\em IEEE Trans. Image Processing}, vol.~13, pp.~600--612, Apr. 2004.
\item P.~Zhang, R. amd~Isola, A.~Efros, E.~Shechtman, and O.~Wang, ``The unreasonable effectiveness of deep features as a perceptual metric,'' {\em IEEE/CVF Conf. Comput. Vis. Pattern Recog.}, pp.~586--595, June 2018.
\end{enumerate}
}

\end{document}